\documentclass{article}

\usepackage[main,preprint]{neurips_2026}

\usepackage[utf8]{inputenc}
\usepackage[T1]{fontenc}
\usepackage{courier}
\usepackage{hyperref}
\usepackage{url}
\usepackage{booktabs}
\usepackage{amsfonts}
\usepackage{amsmath}
\usepackage{amssymb}
\usepackage{mathtools}
\usepackage{nicefrac}
\usepackage{microtype}
\usepackage{xcolor}
\usepackage{graphicx}
\usepackage{array}
\usepackage{enumitem}
\hypersetup{
  pdftitle={PEEK: One-Step-Look-Ahead Exposure Bias Correction for Diffusion Sampling},
  pdfauthor={Yunwei Bai, Ying Kiat Tan, Tsuhan Chen}
}

\newcommand{\method}{PEEK}
\newcommand{\methodg}{PEEK-G}

\newcommand{\eps}{\epsilon}

\title{\method: One-Step-Look-Ahead Exposure Bias Correction for Diffusion Sampling}

\author{
  Yunwei Bai\thanks{Corresponding author.} \quad Ying Kiat Tan \quad Tsuhan Chen\\
  National University of Singapore\\
  baiyunwei@u.nus.edu, yingkiat@u.nus.edu, dprtchen@nus.edu.sg
}

\begin{document}

\maketitle

\begin{abstract}
Diffusion image generators owe much of their practical success
to accurate denoising networks and strong numerical samplers.
However, during sampling,
the denoiser predicts the next step from flawed inputs that
differ substantially from real data.
This train-test gap, or exposure
bias, caps image generation quality (i.e.,
blurry images, missing details, or distorted shapes).
To mitigate the problem, we propose PEEK, a
training-free deterministic sampler plugin.
At each eligible denoising step, PEEK first performs a
provisional one-step
update, evaluates the denoiser once at the look-ahead state,
and interpolates the current clean prediction with this
one-step-later prediction to form the next noisy state.
By taking the additional step, we obtain a generally
more image-like prediction from the higher-signal noisy state,
mitigating exposure bias.
To optimize the interpolation positions, PEEK searches for an
interpolation schedule with a greedy metric-driven search
made efficient by the PEEK
prefix and baseline noise-schedule-rebasing caches.
Across pixel-space and latent-space diffusion models with unconditional, label-guided,
and text-guided settings on CIFAR-10, LSUN Bedroom, ImageNet, and MS-COCO,
the look-ahead correction consistently improves over baselines matched by number of
function evaluations (NFE), with Fr\'echet Inception Distance (FID) reductions up to
49.3\%. The one-time per-model calibration, the only extra cost, 
takes at most 2.2 minutes in our experiments.
Project website is available at \url{https://dachengruoque.github.io/PEEK-Video/}.
\end{abstract}

\section{Introduction}

Diffusion models have become a dominant paradigm for image generation,
combining a supervised denoising objective with an iterative generation
process that gradually transports noise to data \citep{sohl2015deep,
ho2020ddpm,song2021score}. Their practical progress has come from both
sides of the system: more capable denoising models and representations
\citep{dhariwal2021diffusion,rombach2022latent,peebles2023scalable}, and
increasingly accurate samplers that treat generation as the numerical
solution of a reverse-time stochastic or ordinary differential equation
\citep{song2021ddim,karras2022edm,lu2022dpmsolver,lu2022dpmsolverpp,
zhang2023deis,zhao2023unipc}.

\begin{figure*}[t!]
  \centering
  \includegraphics[width=\linewidth]{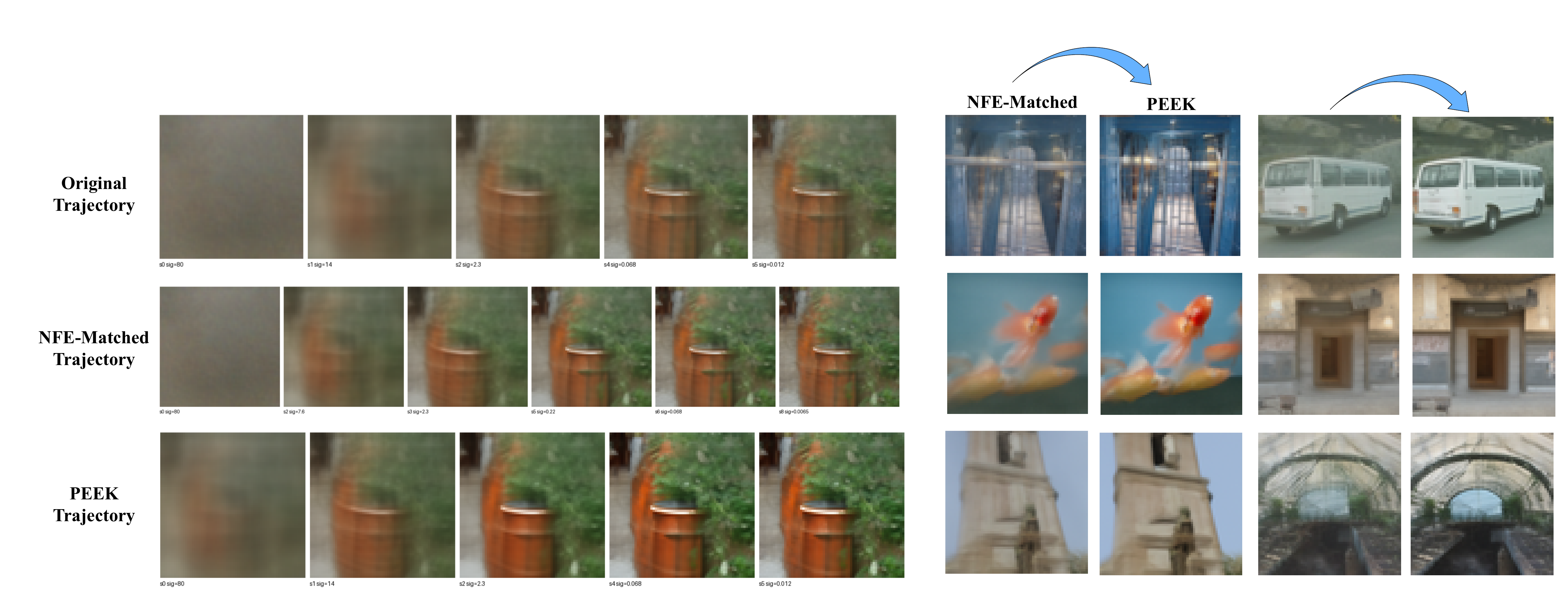}
  \caption{PEEK performs a provisional one-step update,
   and uses the
  look-ahead prediction to interpolate the current prediction at selected
  positions. The selected clean prediction is then used to construct the
  next noisy state. The trajectories illustrate that
  the early-step predictions tend to be especially low-quality, and
   PEEK can yield higher-quality images. Best viewed in color.}
  \label{fig:peek_main}
\end{figure*}

Despite this progress, diffusion sampling remains sensitive to errors
 made along the denoising trajectory, which is most visible
  in the low-number-of-function-evaluations (NFE) regime, where a small
  number of large steps must
  convert an initial high-noise state into a noise-free image.
  Training typically supervises the network on forward-corrupted
  clean data, while inference applies the same network to states
   produced by previous flawed model predictions. The resulting
    states do not follow the same conditional distributions
    encountered during training \citep{ning2023elucidating}.
    Figure~\ref{fig:peek_main} gives
    a visual instance of the PEEK trajectory:
    the denoiser's prediction tends to be low-quality
    especially at the high-noise steps,
    though the training-time counterparts are real high-quality images.
    The final consequences of the exposure bias
    include limited generation quality,
    such as blurry images lacking details of those in the real images.

Existing inference-side methods mostly address the problem indirectly.
Fast ODE solvers reduce discretization error by deriving higher-order
updates \citep{lu2022dpmsolver,lu2022dpmsolverpp}. Exponential-integrator
 and predictor-corrector designs pursue the same low-NFE goal with
 problem-adapted updates
 \citep{zhang2023deis,zhao2023unipc,karras2022edm}.
 Heun-type predictor-corrector samplers, for example, first form an Euler
 predictor and then mix the ODE derivatives estimated at the current and
 predicted states, usually by averaging the two slopes. 
 Schedule optimization methods choose better time grids for a
 fixed sampler \citep{sabour2024align}. Restart sampling adds
 stochastic noise and then re-denoises to contract accumulated
 sampling errors,
 and requires selecting restart intervals and repetition
 counts \citep{xu2023restart}.
 Thus prior training-free improvements primarily change the
 integration rule,
 change the timestep grid, or insert a stochastic re-denoising segment.
 PEEK targets, to our knowledge, a simple,
 \textbf{important} but \textbf{underexplored}
 correction mechanism: \emph{what if we query the denoiser
 at a deterministic provisional
 next state and inject the resulting one-step-later clean prediction
 back to correct exposure bias, instead of just increasing the standard
 inference steps, for image generation quality improvement?}

In short, PEEK is a one-step-look-ahead correction
for the deterministic sampling trajectory.
Given a current noisy state, PEEK computes the usual clean prediction,
takes a provisional next step, obtains the next clean prediction,
and then uses an interpolation gate to form the new noisy
state as input for
the next-step model prediction.
We additionally study a solver-direction variant, \methodg, for
DPM-Solver-style samplers as an extension. Instead of interpolating the
clean prediction used to
form the next noisy state, \methodg\ interpolates the
current solver direction with the
direction implied by the one-step-later prediction.
So far, extensive work has demonstrated that
the generation quality improves along the denoising trajectory,
which is expected
as the later-step predictions are based on the higher signal-to-noise-ratio input.
Furthermore, interchanging the current prediction with the one-step-look-ahead prediction
does not break the inherent logic behind
deterministic diffusion, as different time steps share the same real-image input and
supervision target during training, where only the noise levels differ
\citep{ho2020ddpm,song2021ddim,karras2022edm,
lu2022dpmsolverpp,zhang2023deis,zhao2023unipc}.
Intuitively speaking, assuming perfect corrections
that minimize the train-test gap, the otherwise-drifting
denoising trajectory can
reach the optimal track characterized by only the
empirical training error. In contrast,
if we only densify the denoising trajectory by
increasing the number of ordinary solver
steps, we may continue from the already-drifting trajectory
where the wide early-inference
train-test gaps persist throughout. Therefore, PEEK
has the comparative advantage.

On the other hand, each active interpolation spends an
extra denoiser evaluation and may
exacerbate bias. Therefore,
interpolation positions matter. We avoid manual
hyperparameter tuning through a schedule search, with a
greedy search selecting per-step interpolation gates
using a target metric such as
FID \citep{heusel2017gans}, and the NFE-matched guardrail
rejecting look-ahead
edits that are a weaker use of the same extra evaluation
than an ordinary solver
step. To make this search efficient, we exploit the
cumulative structure of deterministic denoising, where
1) an NFE-matched baseline noise-schedule-rebasing
 cache and 2) a PEEK prefix cache avoid rerunning
 unchanged prefix noise states. Our contributions are:
\begin{itemize}
\item Against the exposure bias problem
in low-NFE deterministic diffusion sampling, we propose
\method, a training-free deterministic one-step-look-ahead
interpolation that edits the current prediction to form
the noisy state for the next-step inference.
\item We extend the same look-ahead principle to \methodg, a
DPM-Solver-compatible solver-direction variant.
\item We introduce a metric-driven interpolation schedule
search for \method(-G), together with baseline noise-schedule-rebasing
and PEEK prefix caches that reduce the 
9-step vanilla search count from 225 to 78 denoiser evaluations,
a 65.3\% reduction.
\item We evaluate PEEK and \methodg\ on pixel-space and
latent-space generation settings, including CIFAR-10
\citep{krizhevsky2009learning},
LSUN Bedroom \citep{yu2015lsun}, ImageNet \citep{deng2009imagenet},
and MS-COCO \citep{lin2014microsoft}, and demonstrate nontrivial
improvements over the NFE-matched baselines, with FID reductions up to 49.3\%,
under matched denoiser-evaluation budgets.
\end{itemize}

\section{Related Work}

\paragraph{Fast Diffusion Sampling.}
Diffusion image generation combines denoising models
\citep{sohl2015deep,ho2020ddpm,song2021score} with numerical samplers.
Low-NFE generation has been improved through deterministic sampling paths,
higher-order ODE solvers, exponential integrators, predictor-corrector updates,
and schedule design \citep{song2021ddim,liu2022pndm,lu2022dpmsolver,
lu2022dpmsolverpp,zhang2023deis,zhao2023unipc,karras2022edm,
sabour2024align}.
These methods mainly change the integration rule or the
timestep grid.
Predictor-corrector samplers instantiate this
idea at different levels: Heun's second-order sampler
takes an Euler predictor, reevaluates the ODE slope
at the predicted endpoint, and applies a trapezoidal
average of the two slopes \citep{karras2022edm},
while UniPC derives unified predictor and corrector
formulas in half-log-SNR space, using current and
previous model outputs so the corrector can
raise solver order without an extra model
evaluation \citep{zhao2023unipc}.
In contrast, PEEK
interpolates the clean-sample prediction with the
one-step-look-ahead counterpart.

\paragraph{Inference Mismatch and Correction.}
Other approaches reduce few-step error by distilling
trajectories or adapting
the model for short sampling \citep{salimans2022progressive,song2023consistency,
oganov2025gas,cai2026mteo}. Training-free inference
corrections are closer to
our setting. Restart sampling inserts forward-noising and
re-denoising segments
to contract accumulated error \citep{xu2023restart}, while SE2P
uses information
from adjacent denoising times in a parallel-processor design
\citep{cisneros2025tango}. PEEK instead spends selected extra
evaluations
on a deterministic one-step look-ahead query and compares against ordinary
NFE-matched trajectories. The appendix gives more related work.

\section{Method}
\subsection{Vanilla \method}
\label{sec:peek}

We describe \method\ for deterministic first-order samplers, the setting
used for the DPM-Solver++, DEIS, and UniPC wrapper experiments
\citep{lu2022dpmsolverpp,zhang2023deis,zhao2023unipc}.
Appendix gives preliminaries. Let
$t_0 > t_1 > \cdots > t_{N-1}$ be decreasing timesteps and $z_i$ the
sampler state at $t_i$. We write the ordinary clean prediction as
\begin{equation}
  \hat{x}_{0,i}^{\mathrm{base}} = D_\theta(z_i,t_i).
\end{equation}
A deterministic solver step is
\begin{equation}
  z_{i+1} = \Phi_i(z_i,\hat{x}_{0,i}^{\mathrm{base}}),
\end{equation}
where $\Phi_i$ includes the sampler-specific conversion from prediction
to the next noisy state.

\paragraph{One-Step Look-Ahead Interpolation.}
At step $i$, \method\ first computes the ordinary clean prediction
$\hat{x}_{0,i}^{\mathrm{base}}$ and an ordinary provisional next state
\begin{equation}
  \tilde{z}_{i+1} = \Phi_i(z_i,\hat{x}_{0,i}^{\mathrm{base}}).
\end{equation}
It then evaluates the denoiser once more at the next solver timestep:
\begin{equation}
  \hat{x}_{0,i}^{\mathrm{look}} = D_\theta(\tilde{z}_{i+1},t_{i+1}).
\end{equation}
The clean prediction is passed through a binary gate:
\begin{equation}
  \hat{x}_{0,i}^{\mathrm{sel}}
  = (1-\gamma_i)\hat{x}_{0,i}^{\mathrm{base}}
  + \gamma_i \hat{x}_{0,i}^{\mathrm{look}},
  \label{eq:peek_interpolate}
\end{equation}
where $\gamma_i \in \{0,1\}$. The final step starts from the original
state $z_i$ but uses the selected clean prediction:
\begin{equation}
  z_{i+1}^{\mathrm{PEEK}} = \Phi_i(z_i,\hat{x}_{0,i}^{\mathrm{sel}}).
  \label{eq:peek_step}
\end{equation}

Let $q_t$ be the forward-noised data marginal and $p_i$ the sampler
marginal at the same timestep. The denoiser is trained under $q_{t_i}$
but queried under $p_i$. If
$\ell_i(x)=\|D_\theta(x,t_i)-m_i(x)\|^2$ is the clean-prediction loss
relative to $m_i(x)=\mathbb{E}_{q}[x_0\mid x_{t_i}=x]$, the trajectory
mismatch induces excess risk
\begin{equation}
  \Delta_i =
  \mathbb{E}_{x\sim p_i}\ell_i(x) -
  \mathbb{E}_{x\sim q_{t_i}}\ell_i(x).
  \label{eq:trajectory_gap}
\end{equation}
Large low-NFE steps can make $\Delta_i$ positive because $p_i$ contains
model-generated states away from the forward-noised data distribution.
Let $x_0^\star$ be a reference clean target and $L_i$ a local Lipschitz
constant of the deterministic step with respect to its clean prediction.
Then
$\|\Phi_i(z,\hat{x}_0)-\Phi_i(z,x_0^\star)\|\le L_i\|\hat{x}_0-x_0^\star\|$,
so reducing clean-prediction error tightens the corresponding upper bound
on one-step state error.

Diffusion training ties predictions at different noise levels to the
same clean target \citep{ho2020ddpm,song2021score,song2021ddim}. The
look-ahead query uses a higher-SNR state while the update remains anchored
at $z_i$, so the correction is local and can propagate through later
deterministic steps.

\subsection{Interpolation Schedule Search}
\label{sec:schedule}

A fully interpolated \method\ can spend unnecessary denoiser calls. We
therefore learn a schedule
$\boldsymbol{\gamma}=(\gamma_0,\ldots,\gamma_{N-2})$
for the $N-1$ peekable gates; the final step has no later noisy timestep.

With candidate set $\mathcal{A}=\{0,1\}$, the search greedily chooses one
gate at a time. Let $M(\boldsymbol{\gamma})$ be the target validation
metric, FID in our experiments. At step $i$, previous gates are fixed and
future gates are zero:
\[
  M_i(\gamma)
  =
  M(\gamma_0,\ldots,\gamma_{i-1},\gamma,0,\ldots,0).
\]
Let $M_i^{\mathrm{base}}$ be the metric of the ordinary solver at the
denoiser-evaluation budget obtained by activating the proposed gate, including
all previously selected nonzero gates. We evaluate each
$\gamma \in \mathcal{A}$ and activate the gate only when it passes both
comparisons:
\begin{equation}
  \gamma_i =
  \begin{cases}
  1, & \text{if }M_i(1)<\min\{M_i(0),M_i^{\mathrm{base}}\},\\
  0, & \text{otherwise}.
  \end{cases}
\end{equation}
The guardrail therefore compares an active correction with spending its extra
evaluation on an ordinary solver step. The zero candidate retains the current
partial schedule at its lower budget.

For each candidate position, the search reuses the selected prefix state,
evaluates the NFE-matched baseline, evaluates the current partial schedule,
and tests each candidate gate. The selected state becomes the next prefix.
Appendix gives pseudocode and a real search-log excerpt.

A nonzero $\gamma_i$ means the active candidate outperforms both the current
partial schedule and the NFE-matched ordinary baseline. A zero means the
active candidate fails at least one of these two comparisons.

The search is a one-time calibration for a model, sampler, step count,
and target metric. At deployment, the schedule is fixed. A \method\ sample
with $N$ solver steps and $K$ nonzero gates uses $N+K$ denoiser
evaluations, and our main comparisons use ordinary baselines with the same
budget.

\subsection{PEEK-G Extension}
\methodg\ forms the same provisional one-step-look-ahead state as \method,
and blends the current and look-ahead solver directions.
The appendix gives the formal
DPM-Solver direction notation.
\subsection{Search-Time Cache Optimization}
\label{sec:cache}

\begin{figure*}[t]
  \centering
  \includegraphics[width=\linewidth]{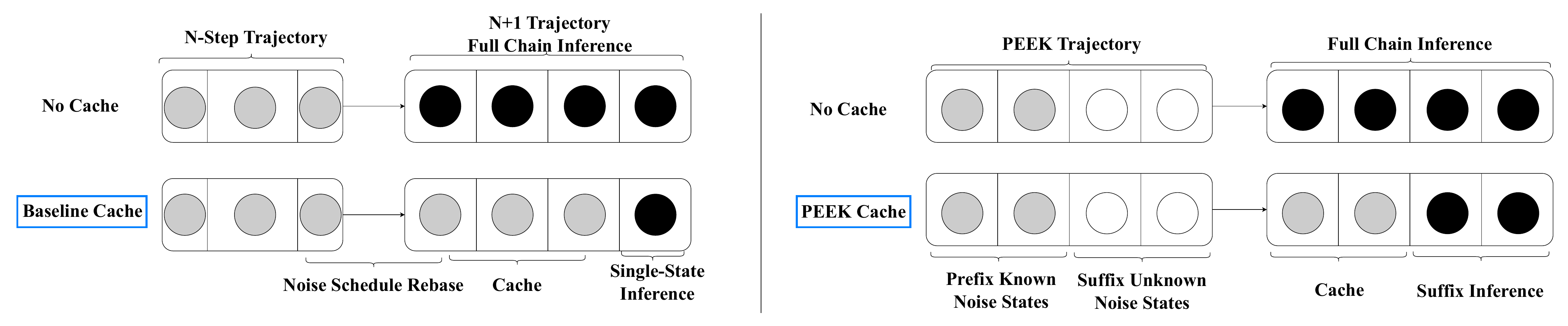}
  \caption{Caches used during schedule search. Baseline rebasing avoids
  rerunning ordinary prefixes as the NFE-matched chain grows,
  while the \method\ prefix cache lets later candidates run only the
  remaining suffix. Best viewed in color.}
  \label{fig:cache}
\end{figure*}

\method\ uses two caches for deterministic cumulative first-order search.
Figure~\ref{fig:cache} summarizes their complementary reuse paths.

\paragraph{Baseline Noise-Schedule Rebasing Cache.}
For a deterministic sampler state $z_t$ and clean prediction $\hat{x}_0$, the
variance-preserving implied noise is
\begin{equation}
  \hat{\eps} = \frac{z_t-\sqrt{\bar{\alpha}_t}\hat{x}_0}{\sqrt{1-\bar{\alpha}_t}}.
  \label{eq:implied_noise}
\end{equation}
For target timestep $s$, the rebased state is
\begin{equation}
  z_s = \sqrt{\bar{\alpha}_s}\hat{x}_0 + \sqrt{1-\bar{\alpha}_s}\hat{\eps}.
  \label{eq:rebase}
\end{equation}
Equation~\eqref{eq:rebase} moves cached ordinary prefixes to the next
NFE-matched grid without recomputing them. Substituting
Eq.~\eqref{eq:implied_noise} into Eq.~\eqref{eq:rebase} gives
\begin{align}
  R_{t\rightarrow s}(z_t,\hat{x}_0)
  &= \sqrt{\bar{\alpha}_s}\hat{x}_0
  + \sqrt{\frac{1-\bar{\alpha}_s}{1-\bar{\alpha}_t}}
  \left(z_t-\sqrt{\bar{\alpha}_t}\hat{x}_0\right) \nonumber\\
  &= c_x z_t + c_0 \hat{x}_0,
  \label{eq:rebase_operator}
\end{align}
where $c_x=\sqrt{(1-\bar{\alpha}_s)/(1-\bar{\alpha}_t)}$
and $c_0=\sqrt{\bar{\alpha}_s}-c_x\sqrt{\bar{\alpha}_t}$.
Thus a cached pair $(z_t,\hat{x}_0)$ gives the target-timestep state
$z_s=R_{t\rightarrow s}(z_t,\hat{x}_0)$. The baseline cache is an
approximation because step-interval differences can cause small mismatch.

\paragraph{\method\ Prefix Cache.}
After selecting $\gamma_i$, the state at $t_{i+1}$ is independent of
future gates. We cache it, so a later candidate at position $j$ starts from
the cached state after step $j-1$ and computes only the suffix.

\paragraph{Search Cost.}
Let candidate index $i=0,\ldots,N-2$, so its NFE-matched budget is
$N+i+1$ in the worst case where every preceding gate is nonzero.
Without caching, the initial partial schedule costs $N$ queries, and both
the ordinary baseline and the \method\ candidate are rerun at every budget:
\begin{equation}
  C_n
  = N + 2\sum_{i=0}^{N-2}(N+i+1)
  = 3N^2-2N.
\end{equation}
With caching, the initial partial schedule is evaluated once at a cost of
$N$ queries. The first $N+1$-query ordinary baseline is then evaluated in
full. At each later position $i=1,\ldots,N-2$, rebasing converts the
cached pre-final state to the new grid, leaving only the new final-step
query. Thus, the baseline cost is
\begin{equation}
  C_b
  = N+1+\sum_{i=1}^{N-2}1
  = 2N-1.
\end{equation}
For the \method\ candidate at position $i$, its cached prefix is free;
one new base query, one new look-ahead query, and $N-1-i$ ordinary suffix
queries remain. Hence
\begin{equation}
  C_p
  = \sum_{i=0}^{N-2}\left[2+(N-1-i)\right]
  = \frac{N^2+3N-4}{2}.
\end{equation}
Including the initial partial schedule, the cached total is therefore
$C_c=N+C_b+C_p=(N^2+9N-6)/2$. For $N=9$, the naive and cached counts are
225 and 78, respectively, a 65.3\%
reduction. Appendix~\ref{app:cache_details} gives the expanded count.

\section{Experiments}
\label{sec:experiments}

\begin{figure*}[t]
  \centering
  \includegraphics[width=\linewidth]{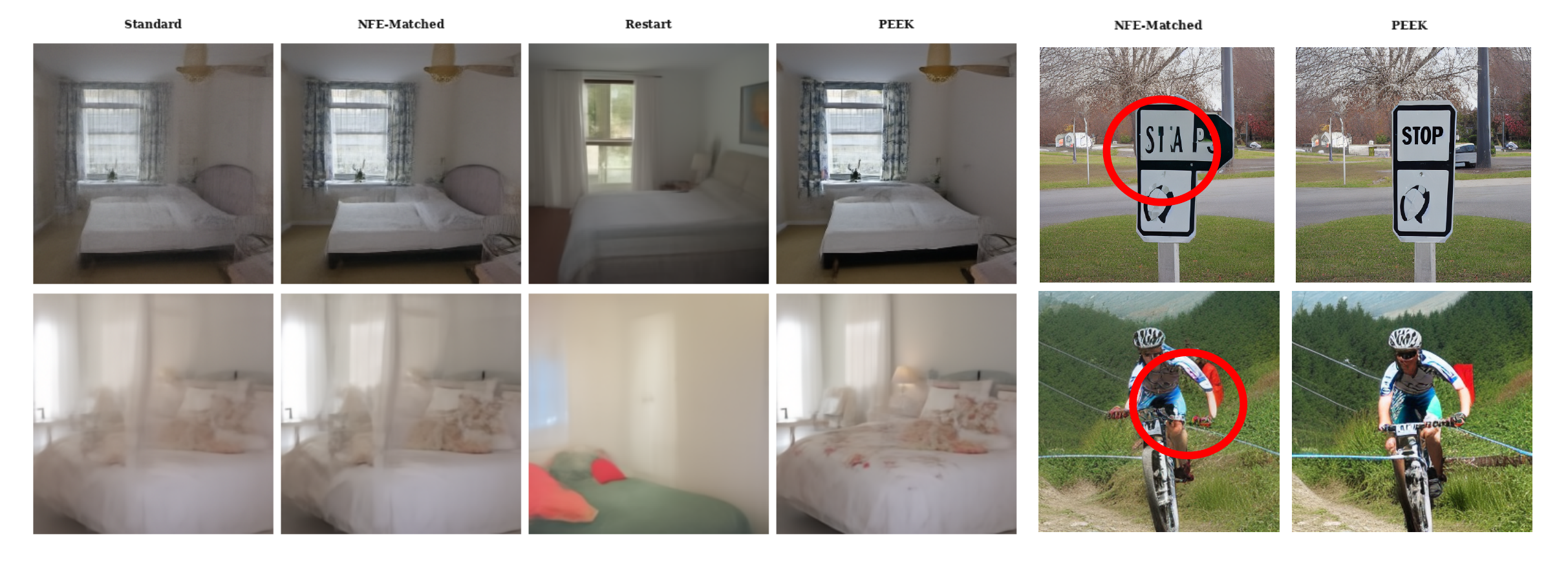}
  \caption{Qualitative Comparison. \method\ uses clean-prediction interpolation and
  generally achieves better detail-rendering or shape coherence for the generated images. Best viewed in color.}
  \label{fig:qual_compare}
\end{figure*}

\begin{figure*}[t]
  \centering
  \includegraphics[width=\linewidth]{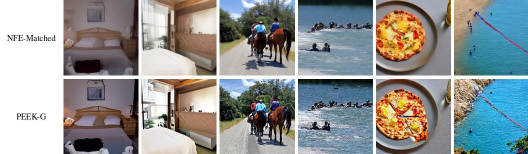}
  \caption{\methodg\ Qualitative Comparison. \methodg\ uses solver-direction interpolation and
  generally achieves better detail-rendering or shape coherence for the generated images.
  Best viewed in color.}
  \label{fig:peekgrad_good_main}
\end{figure*}

\begin{figure*}[t]
  \centering
  \includegraphics[width=0.48\linewidth]{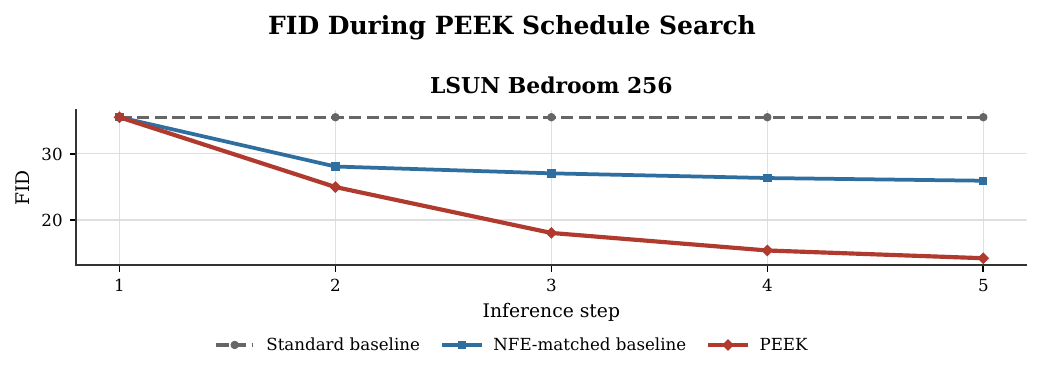}
  \hfill
  \includegraphics[width=0.48\linewidth]{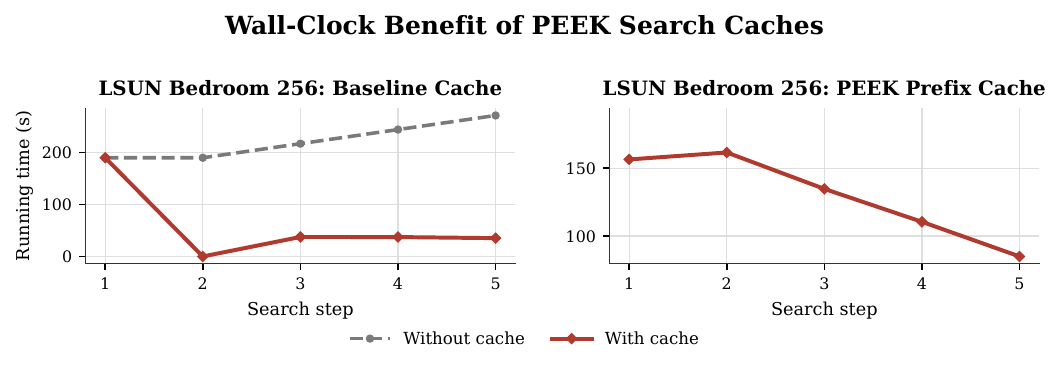}
  \caption{Schedule Search and Cache Behavior.
  The \method\ curve improves as useful interpolation positions
  are accepted.
  Prefix reuse and baseline rebasing reduce repeated denoising work
  during schedule search, especially as the selection
  moves toward later timesteps.}
  \label{fig:search_cache_lines}
\end{figure*}

\begin{table*}[t]
  \centering
  \small
  \setlength{\tabcolsep}{3.5pt}
  \begin{tabular}{@{}lccc@{}}
    \toprule
    Method & \shortstack{LSUN FID\\(Base NFE 6) $\downarrow$} & Recall $\uparrow$ & Coverage $\uparrow$ \\
    \midrule
    DEIS, NFE-Matched & 20.57 & 0.521 & 0.558 \\
    DPM-Solver++, NFE-Matched & 15.95 & 0.567 & 0.656 \\
    UniPC, NFE-Matched & 12.11 & 0.601 & 0.773 \\
    Heun's 2nd, NFE-Matched & 51.69 & 0.369 & 0.254 \\
    Restart-style repeat & 30.97 & 0.330 & 0.400 \\
    \midrule
    \method, DEIS & \textbf{11.38} (44.7\%)$\downarrow$ & 0.614 & 0.744 \\
    \method, DPM-Solver++ & 11.59 (27.3\%)$\downarrow$ & \textbf{0.627} & \textbf{0.755} \\
    \bottomrule
  \end{tabular}
  \caption{\method\ main results on the basic LSUN model \citep{yu2015lsun}. Lower FID is better, while higher recall and coverage are better.}
  \label{tab:main_peek}
\end{table*}

\begin{table*}[t]
  \centering
  \small
  \setlength{\tabcolsep}{4pt}
  \begin{tabular}{@{}lccc@{}}
    \toprule
    \multicolumn{4}{c}{(a) Additional Dataset FID} \\
    \midrule
    Dataset & Base NFE & NFE-Matched FID $\downarrow$ & \method\ FID $\downarrow$ \\
    \midrule
    CIFAR-10 $32 \times 32$ (Model A, 9 steps) & 9 & 25.80 & \textbf{20.23} (21.6\%)$\downarrow$ \\
    CIFAR-10 $32 \times 32$ (Model A, 20 steps) & 20 & 18.16 & \textbf{15.05} (17.1\%)$\downarrow$ \\
    CIFAR-10 $32 \times 32$ (Model B, 9 steps) & 9 & 19.87 & \textbf{16.46} (17.2\%)$\downarrow$ \\
    CIFAR-10 $32 \times 32$ (Model B, 20 steps) & 20 & 12.23 & \textbf{10.04} (17.9\%)$\downarrow$ \\
    ImageNet $256 \times 256$ & 9 & 20.40 & \textbf{19.64} \\
    MS-COCO $512 \times 512$ (order 1) & 9 & 14.58 & \textbf{12.93} (11.3\%)$\downarrow$ \\
    MS-COCO $512 \times 512$ (order 3) & 9 & 13.54 & \textbf{11.74} (13.3\%)$\downarrow$ \\
    \bottomrule
  \end{tabular}
  \par\smallskip
  \begin{tabular}{@{}lcccc@{}}
    \toprule
    \multicolumn{5}{c}{(b) Proxy Search} \\
    \midrule
    Dataset & Base NFE & \shortstack{256-Dim Proxy\\FID $\downarrow$} & \shortstack{Search Time /\\Peak CUDA Memory} & Final FID $\downarrow$ \\
    \midrule
    ImageNet $256 \times 256$ & 9 & 5.15 & 1.5 min / 39.0 GB & 19.64 \\
    MS-COCO $512 \times 512$ & 9 & 2.79 & 2.2 min / 102.2 GB & 12.93 \\
    \bottomrule
  \end{tabular}
  \caption{Additional datasets and proxy-search results with \method\ and DPM-Solver++ \citep{lu2022dpmsolverpp}.}
  \label{tab:extra}
\end{table*}

\begin{table*}[t]
  \centering
  \small
  \setlength{\tabcolsep}{4pt}
  \begin{tabular}{@{}lllccc@{}}
    \toprule
    Dataset / model & Resolution & Solver & Base NFE & NFE-Matched FID $\downarrow$ & \method\ FID $\downarrow$ \\
    \midrule
    ImageNet-EDM & $64 \times 64$ & DPM-Solver++  & 6 & 19.71 & \textbf{9.99} (49.3\%)$\downarrow$ \\
    ImageNet-EDM & $64 \times 64$ & DEIS  & 6 & 10.13 & \textbf{7.29} (28.0\%)$\downarrow$ \\
    \midrule
    LSUN-Bedroom-EDM & $256 \times 256$ & DPM-Solver++  & 5 & 22.89 & \textbf{11.77} (48.6\%)$\downarrow$ \\
    LSUN-Bedroom-EDM & $256 \times 256$ & DEIS  & 5 & 12.10 & \textbf{8.79} (27.4\%)$\downarrow$ \\
    LSUN-Bedroom-EDM & $256 \times 256$ & iPNDM  & 5 & 15.86 & \textbf{10.69} (32.6\%)$\downarrow$ \\
    \bottomrule
  \end{tabular}
  \caption{\method\ vs the NFE-matched baselines on EDM-trained models.}
  \label{tab:edm_results}
\end{table*}

\begin{table}[t]
  \centering
  \small
  \setlength{\tabcolsep}{4pt}
  \begin{tabular}{@{}lccc@{}}
    \toprule
    Dataset  & Base NFE & NFE-Matched FID $\downarrow$ & \methodg\ FID $\downarrow$ \\
    \midrule
    CIFAR-10 (Model B)  & 9 & 17.48 & \textbf{13.18} (24.6\%)$\downarrow$ \\
    Basic LSUN model  & 6 & 15.95 & \textbf{15.60} \\
    ImageNet  & 9 & 20.40 & \textbf{19.68} \\
    MS-COCO  & 9 & 14.41 & \textbf{10.66} (26.0\%)$\downarrow$ \\
    \bottomrule
  \end{tabular}
  \caption{\methodg\ versus the NFE-matched DPM-Solver++ baseline.}
  \label{tab:peekg_results}
\end{table}

\begin{table}[ht]
  \centering
  \small
  \setlength{\tabcolsep}{4pt}
  \begin{tabular}{lccc}
    \toprule
    Method/$\boldsymbol{\gamma}$ & NFE & DPM-Solver++ & DEIS \\
    \midrule
    \method{} 01111 (Greedy Search Schedule) & 10 & \textbf{11.59} & \textbf{11.38} \\
    \method{} 10111 & 10 & 12.63 & 13.34 \\
    \method{} 11011 & 10 & 13.82 & 14.39 \\
    \method{} 11101 & 10 & 13.50 & 13.51 \\
    \method{} 11110 & 10 & 13.90 & 13.36 \\
    \midrule
    NFE-Matched Baseline & 11 & 14.43 & 18.45 \\
    \method{} 11111 & 11 & 12.89 & 12.98 \\
    \bottomrule
  \end{tabular}
  \caption{Basic LSUN Bedroom model schedule ablation.}
  \label{tab:lsun_schedule_ablation}
  \par\medskip
  \setlength{\tabcolsep}{2pt}
  \begin{tabular}{@{}lcccc@{}}
    \toprule
    Dataset / model & Base NFE & Total NFE & NFE-Matched FID-5K $\downarrow$ & All-One \method\ FID-5K $\downarrow$ \\
    \midrule
    CIFAR-10 (Model A) & 9 & 17 & 34.21 & \textbf{29.15} (14.8\%)$\downarrow$ \\
    ImageNet-EDM $64 \times 64$ & 6 & 11 & 20.93 & \textbf{15.05} (28.1\%)$\downarrow$ \\
    LSUN-Bedroom-EDM $256 \times 256$ & 5 & 9 & 21.59 & \textbf{15.50} (28.2\%)$\downarrow$ \\
    \bottomrule
  \end{tabular}
  \caption{Fixed all-one \method\ schedules without schedule search.}
  \label{tab:allone_results}
\end{table}

\paragraph{Evaluation Protocol.}
We evaluate unconditional pixel-space generation on CIFAR-10
\citep{krizhevsky2009learning} and LSUN Bedroom \citep{yu2015lsun},
class-conditional pixel-space generation on ImageNet at $64 \times 64$,
class-conditional latent generation on ImageNet at $256 \times 256$
\citep{deng2009imagenet},
 and text-conditioned latent generation on MS-COCO \citep{lin2014microsoft}.
 For CIFAR-10 (Models A and B), LSUN Bedroom (basic and EDM), and ImageNet
 ($64 \times 64$ and $256 \times 256$), we evaluate two models per dataset.
  We report Fr\'echet Inception Distance (FID) \citep{heusel2017gans},
    where lower is better. All methods in a comparison use the same
    model and evaluation protocol unless stated otherwise.
Parenthetical percentages in the result tables report relative FID reductions
over the corresponding NFE-matched baseline and are shown only when they exceed 5\%.

\paragraph{Baselines.}
 The NFE-matched baseline uses the same solver
  with additional ordinary steps so that it spends
  the same number of denoiser evaluations as \method. 
  The NFE-matched baseline uses the same default schedule 
  as PEEK's to ensure fairness. 
   We compare DPM-Solver++, DEIS, Heun's 2nd and UniPC through
    the solver wrappers used by
     \citet{zhou2023fast,chen2024trajectory}. Heun's 2nd and UniPC are included
     as explicit predictor-corrector baselines. We also
      include a Restart-style repeat
      baseline in Restart sampling \citep{xu2023restart}.
      Note that the original Restart sampling requires
      tuning of intervals and denoising repetition steps, but
      the exact method of tuning
      is underspecified. Therefore, we compare the Restart correction
      mechanism against
      ours using the exact same configurations (e.g., NFE), so as to
      keep variables
      controlled and contrast the core replacement mechanisms directly.
      The appendix specifies this baseline in more detail.

\paragraph{Models and Compute.}
For CIFAR-10 and the basic LSUN Bedroom model we use pretrained DDPM-family
pixel-space models \citep{ho2020ddpm}. CIFAR-10 is evaluated at $32 \times 32$;
LSUN Bedroom at $256 \times 256$; ImageNet at $64 \times 64$ or
$256 \times 256$, depending on the model; and MS-COCO at $512 \times 512$.
For ImageNet we use a
DiT-XL/2-256 latent diffusion model with classifier-free guidance
\citep{peebles2023scalable,ho2022classifierfree}. For MS-COCO we
use a Stable Diffusion v1.5 latent model with the conventional
text-guided evaluation configuration \citep{rombach2022latent}.
We additionally evaluate ImageNet-EDM ($64 \times 64$) and the EDM LSUN model ($256 \times 256$), both
based on the EDM training paradigm \citep{karras2022edm}; for the
EDM LSUN model, we include iPNDM \citep{liu2022pndm}.
Schedule search and final sampling were run on multi-GPU H200 nodes.
Pixel-space \method\ applies the interpolation in image space;
latent-space \method\ applies the same clean-prediction
interpolation in latent space before decoding for evaluation.
Unless otherwise stated, all \method\ and \methodg\ visual samples use DPM-Solver++
schedules and sampling. In schedule search, vanilla \method\ uses gate candidates
\(\{0,1\}\), while \methodg\ uses \(\{0,0.25,0.5,0.75,1\}\).

\paragraph{\method\ Results.}

Table~\ref{tab:main_peek} shows that \method\ improves over
strong NFE-matched baselines. With DEIS, FID decreases from
20.57 to 11.38 on the basic LSUN model,
a 44.7\% improvement. With DPM-Solver++, \method\ decreases FID
from 15.95 to 11.59, a 27.3\% improvement.
The Restart-style repeat baseline under the same NFE budget
is worse in these runs, suggesting our comparative advantage
in the core exposure bias correction mechanism.
Both \method\ and Restart-style repeat spend
extra evaluations on correction rather than model
retraining, but the correction
signal is different: \method\ remains on the deterministic
trajectory and edits
the current clean prediction, while the Restart-style baseline
revisits a noised state
for stochastic re-noising and denoising.
Heun's 2nd performs worse than the DPM-Solver++, DEIS, and UniPC
baselines
in these low-NFE settings, aligning with related reports and
demonstrating our
comparative advantage over this traditional predictor-corrector
mechanism
\citep{wang2025pas}.

The NFE-matched rows are the intentionally compute-matched baselines.
Improvements over these rows therefore indicate that \method\ is improving
the use of computation.
The PRDC results support the FID comparison. Relative to their NFE-matched
baselines, DEIS \method\ raises recall from 0.521 to 0.614 and coverage from
0.558 to 0.744, while DPM-Solver++ \method\ raises recall from 0.567 to 0.627
and coverage from 0.656 to 0.755.

Table~\ref{tab:extra} reports additional datasets.
On CIFAR-10, schedules are selected with the
proxy FID and then rerun for standard final FID.
To assess the sensitivity of \method\ to model choice,
we test CIFAR-10 Models A and B, which are trained
on the same dataset but have different generation quality.
With CIFAR-10 Model A, a 9-step base solver
reduces FID from 25.80 to 20.23, a 21.6\% improvement. Increasing the base
solver to 20 steps lowers both FID values, moving the NFE-matched baseline to
18.16 and \method\ to 15.05, a 17.1\% improvement. With CIFAR-10 Model B,
the same study gives
lower absolute FID values: \method\ reduces 9-step FID from
$19.87$ to $16.46$ (17.2\%), and 20-step FID from
$12.23$ to $10.04$ (17.9\%).

On ImageNet,
the binary proxy-selected \method\ schedule reduces FID from
$20.40$ to $19.64$. On MS-COCO, the same proxy-search
protocol reduces first-order DPM-Solver++ FID from $14.58$ to
$12.93$, an 11.3\% relative reduction. These results suggest that one-step-look-ahead correction can
apply beyond one resolution, architecture family, or conditioning mode, but the
effectiveness varies across models and variants.
Note that the ImageNet result tests a class-conditional latent transformer,
while MS-COCO tests text-guided latent sampling with a different
prompt-conditioned pipeline.

Table~\ref{tab:edm_results} adds checkpoints trained under
the EDM paradigm \citep{karras2022edm}. On ImageNet-EDM at $64 \times 64$,
\method\ reduces FID by 49.3\% for first-order DPM-Solver++ and 28.0\% for
first-order DEIS under the same compute. On the EDM LSUN model, the
corresponding improvements are 48.6\% for DPM-Solver++, 27.4\% for DEIS,
and 32.6\% for iPNDM \citep{liu2022pndm}. These EDM LSUN rows use a
different EDM-trained model from the basic LSUN model in Table~\ref{tab:main_peek},
so the two LSUN blocks should be read as separate models.

\paragraph{Higher-Order Solver.}
We also evaluate
third-order DPM-Solver++ on MS-COCO with the same binary interpolation schedule.
The higher-order NFE-matched baseline improves over the first-order baseline
from $14.58$ to $13.54$ FID, and \method\ further
improves the higher-order run to $11.74$ FID, a 13.3\% relative reduction
over its NFE-matched baseline, suggesting compatibility.

\paragraph{Schedule Effectiveness.}
Figure~\ref{fig:search_cache_lines} shows representative
 search traces.
 The selected schedules are not uniformly dense.
 Interpolating at all peekable positions does not worsen the ordinary prediction
 on the basic LSUN Bedroom model. However, not every extra denoiser call is equally valuable,
 and very late interpolations can have limited impact because much of the layout
 has already been determined, while the early extra denoiser call may remain low-quality
 and insufficient to beat the NFE-matched baseline at one extra inference step.
 Therefore, the early extra denoiser call may also be skipped.
 Table~\ref{tab:lsun_schedule_ablation} tests this directly on the basic LSUN Bedroom model
 by holding the number of interpolations fixed and changing only which candidate
 position is disabled; binary strings list the five peekable positions from
 left to right. It also reports the all-one schedule and its 11-step ordinary
baseline. In this single-evaluation ablation, the all-one schedule improves
its baseline but yields a higher FID than the searched schedule.
All-one interpolation improves over its own NFE-matched 11-step baseline by 10.7\%
for DPM-Solver++ and 29.6\% for DEIS. The searched schedule also uses fewer
interpolation calls than the all-one schedule.

Table~\ref{tab:allone_results} evaluates the fixed all-one schedule more broadly,
without schedule search, on DPM-Solver++. It improves over the NFE-matched baseline by 
28.1\% on ImageNet-EDM,
and 28.2\% on the EDM LSUN model. The all-one schedule requires no
search and stills provide non-trivial generation-quality gains, 
demonstrating PEEK effectiveness at almost zero overhead. 
Nevertheless, the schedule search remains
useful for selecting a sparser, higher-quality schedule.

\paragraph{Cache Effectiveness.}
Figure~\ref{fig:search_cache_lines} summarizes the
runtime effect of the cache strategy. The main gain comes from
avoiding repeated prefixes: once the search has accepted the first
several interpolation gates, testing a later candidate only requires
denoising the remaining suffix. 

\paragraph{PEEK-G Extension.}
Table~\ref{tab:peekg_results} reports \methodg\ results on CIFAR-10 Model B,
the basic LSUN model, ImageNet, and MS-COCO.
It reduces FID by 24.6\% on CIFAR-10 Model B and 26.0\% on MS-COCO;
the corresponding reductions on the basic LSUN model and ImageNet are
2.2\% and 3.5\%, respectively.

\paragraph{Qualitative Comparison.}
Figure~\ref{fig:qual_compare} compares standard,
NFE-matched, Restart-style repeat, and \method\ samples.
\method\ tends to preserve the
deterministic trajectory while sharpening object details.
In contrast, Restart-style repeat can introduce larger stochastic
deviations because it explicitly adds fresh noise before denoising.
Figure~\ref{fig:peekgrad_good_main} compares the NFE-matched samples against
those generated by \methodg, which renders clearer details.
The appendix includes additional qualitative samples.

\section{Conclusion}

We presented \method, a training-free and model-agnostic algorithm for improving
deterministic diffusion sampling.
  By interpolating the current clean prediction with its look-ahead
  counterpart at searched positions to construct the next noisy state,
  \method\ improves the NFE-matched baselines in diverse
  image generation settings.
  The cache design makes schedule calibration more efficient.
  A current limitation is the drop in coverage,
  which could be mitigated as future work.
  Overall,
  \method\ spends inference computation on generally cleaner signals,
  potentially improving generation across different models and samplers, 
  without model fine-tuning and or deployment overhead beyond the per-model calibration cost.

\clearpage
\section*{Funding Disclosure}
Funding in direct support of this work was provided in part by the
Ministry of Education, Singapore, under its MOE AcRF Tier 3 Grant
(MOE-MOET32022-0001). The authors declare no competing interests.
\bibliographystyle{plainnat}
\bibliography{references}

\clearpage
\appendix
\section{Diffusion and Sampling Preliminaries}
\label{app:diffusion_prelim}

This appendix briefly reviews the diffusion notation used in the main text
and clarifies how PEEK fits into standard deterministic sampling. We focus
on the variance-preserving notation used in our cache derivation and on the clean-prediction
interface used by the samplers in our experiments.

\paragraph{Forward Noising Process.}
Diffusion models define a family of noisy marginals by gradually corrupting a
clean data sample \(x_0\). Under the standard variance-preserving parameterization,
the noisy variable at diffusion time \(t\) can be written as
\[
x_t = \sqrt{\bar{\alpha}_t} x_0 + \sqrt{1-\bar{\alpha}_t}\epsilon,
\qquad
\epsilon \sim \mathcal{N}(0,I),
\]
or equivalently
\[
q(x_t \mid x_0)
=
\mathcal{N}\!\left(
\sqrt{\bar{\alpha}_t}x_0,
(1-\bar{\alpha}_t)I
\right).
\]
Here \(\bar{\alpha}_t\) is the cumulative signal coefficient. Large diffusion times
correspond to high-noise states, where \(\bar{\alpha}_t\) is small and most clean-image
structure has been obscured. Small diffusion times correspond to low-noise states, where
the clean image is more visible.

The marginal distribution of forward-noised real data at time \(t\) is denoted by \(q_t\).
During training, the denoising network is supervised on samples \(x_t \sim q_t\), so its inputs
are noisy versions of real data rather than states produced by its own previous predictions.

\paragraph{Clean-Prediction Notation.}
Denoising networks can be parameterized in several equivalent ways, such as predicting noise,
predicting the clean sample, or predicting a velocity variable. PEEK operates on the clean prediction,
so throughout the paper we write
\[
\hat{x}_{0,i} = D_\theta(z_i,t_i)
\]
for the clean-sample estimate obtained from the denoiser at sampler state \(z_i\) and timestep
\(t_i\). If the underlying model predicts noise \(\epsilon_\theta(x_t,t)\), this clean estimate
is obtained by the usual conversion
\[
D_\theta(x_t,t)
=
\frac{x_t-\sqrt{1-\bar{\alpha}_t}\epsilon_\theta(x_t,t)}
{\sqrt{\bar{\alpha}_t}}.
\]
Other prediction parameterizations are converted analogously. Thus \(D_\theta\) should be
read as the denoiser after conversion to a clean-sample estimate, not necessarily as the model's native output.

The subscript convention is important. The subscript \(0\) in \(x_0\) and \(\hat{x}_{0,i}\)
denotes the clean endpoint of the diffusion process. We use \(z_i\) for the noisy sampler
state at discrete index \(i\) and timestep \(t_i\); \(\hat{x}_{0,i}\) is the clean prediction
made from that state. This distinction avoids overloading \(x_0\) as both the clean endpoint
and the initial high-noise sampler state.

\paragraph{Deterministic Sampling.}
At inference time, a deterministic sampler starts from a high-noise state and follows a
decreasing timestep grid
\[
t_0 > t_1 > \cdots > t_{N-1}.
\]
Let \(z_i\) denote the sampler state at timestep \(t_i\). A first-order deterministic
solver step can be written abstractly as
\[
z_{i+1} = \Phi_i(z_i,\hat{x}_{0,i}),
\qquad
\hat{x}_{0,i}=D_\theta(z_i,t_i),
\]
where \(\Phi_i\) includes the sampler-specific conversion between the clean prediction,
the implied noise prediction, and the next noisy state. This notation covers the deterministic
first-order wrappers used in our experiments.

For example, a DDIM-style update under the same variance-preserving parameterization first
computes the implied noise
\[
\hat{\epsilon}_i
=
\frac{z_i-\sqrt{\bar{\alpha}_{t_i}}\hat{x}_{0,i}}
{\sqrt{1-\bar{\alpha}_{t_i}}},
\]
and then forms the next state by reusing this clean/noise decomposition at the lower-noise timestep:
\[
z_{i+1}
=
\sqrt{\bar{\alpha}_{t_{i+1}}}\hat{x}_{0,i}
+
\sqrt{1-\bar{\alpha}_{t_{i+1}}}\hat{\epsilon}_i.
\]
Different deterministic solvers use different numerical updates, but they share the same
practical structure: the denoiser is queried at the current state, the prediction is converted
into the sampler's preferred parameterization, and the next state is produced without adding
fresh random noise.

\paragraph{DPM-Solver Direction Form.}
\label{app:dpmsolver_prelim}
DPM-Solver-style samplers write the reverse diffusion dynamics as an ODE in a noise-time
variable and approximate the integral of a model-derived direction over each solver interval
\citep{lu2022dpmsolver,lu2022dpmsolverpp}. Abstractly, the update can be written as
\[
z_{i+1}
= A_i z_i + \sum_{j=0}^{r-1} B_{i,j} d_{i-j},
\qquad
d_i = G_i(z_i,D_\theta(z_i,t_i)),
\]
where the coefficients \(A_i\) and \(B_{i,j}\) are determined by the timestep grid and solver
order, and \(d_i\) is the current model direction after converting the denoiser output into the
solver's parameterization. First-order DPM-Solver++ uses only the current direction; multistep
variants additionally use stored directions from previous timesteps.

\subsection{\methodg\ Formalization}
\label{app:peekg_formalization}
\method\ interpolates the clean prediction before the next noisy state is
formed. For DPM-Solver-style updates, the same look-ahead information can
also be used at the solver-direction level. Let
\begin{equation}
  \hat{x}_{0,i}^{\mathrm{base}} = D_\theta(z_i,t_i),
  \qquad
  d_i^{\mathrm{base}} = G_i(z_i,\hat{x}_{0,i}^{\mathrm{base}})
\end{equation}
denote the model direction used by the deterministic solver at timestep
\(t_i\), where \(G_i\) is the sampler-specific conversion from the denoiser
output to the ODE direction. At a nonzero gate, \methodg\ first forms the same
provisional next state \(\tilde{z}_{i+1}\) as in \method\ using the ordinary
scheduler step, but does not commit that state. It then evaluates the denoiser
at the look-ahead state,
\begin{equation}
  \hat{x}_{0,i}^{\mathrm{look}}
  = D_\theta(\tilde{z}_{i+1},t_{i+1}),
  \qquad
  d_i^{\mathrm{look}}
  = G_{i+1}(\tilde{z}_{i+1},\hat{x}_{0,i}^{\mathrm{look}}).
\end{equation}
For a nonzero gate, the selected solver direction is
\begin{equation}
  d_i^{\mathrm{sel}}
  = (1-\gamma_i)d_i^{\mathrm{base}}+\gamma_i d_i^{\mathrm{look}}.
  \label{eq:peekgrad_interpolate}
\end{equation}
For a zero gate, \methodg\ instead sets
\(d_i^{\mathrm{sel}}=d_i^{\mathrm{base}}\) without computing
\(d_i^{\mathrm{look}}\).
The actual DPM-Solver step is still applied from the current state \(z_i\)
with the usual timestep coefficients for the current solver interval.
Fractional gates are natural in this direction space, so our \methodg\ searches use
\(\gamma_i\in\{0,0.25,0.5,0.75,1\}\). The solver coefficients remain those of
the current interval, so \methodg\ is not a new DPM-Solver derivation. It is a
post-hoc direction correction using a model prediction from a higher-SNR state.
This is mathematically coherent with the diffusion target because denoiser
predictions at different timesteps are trained toward the same clean data
endpoint \(x_0\), after the appropriate parameterization conversion. The
practical risk is that the look-ahead prediction may be confidently biased; in
that case \methodg\ can amplify the bias even while sharpening the local
direction.

\paragraph{NFE Accounting.}
We count one function evaluation for each call to \(D_\theta\). A base deterministic
trajectory with \(N\) denoiser queries therefore uses \(N\) NFEs. PEEK can be applied at the first
\(N-1\) positions, because the final denoising query has no later noisy timestep inside the trajectory
from which to obtain a look-ahead prediction.

If \(K\) interpolation gates in the PEEK schedule are nonzero, then PEEK makes \(K\) additional
look-ahead denoiser calls. Its deployment cost is therefore \(N+K\) NFEs. The NFE-matched baseline
uses the same solver with \(N+K\) ordinary denoiser evaluations and no look-ahead interpolation.

\paragraph{Training Distribution versus Sampling Trajectory.}
A useful way to view low-NFE sampling is through the mismatch between the training inputs and
the inference-time states. During training, the denoiser is optimized on \(x_t \sim q_t\), where
\(q_t\) is obtained by forward noising real data. During deterministic sampling, however, the state
\(z_i\) is generated recursively by the sampler and the model's own previous predictions.
Let \(p_i\) denote the marginal distribution of these generated states at timestep \(t_i\). In general,
\[
p_i \neq q_{t_i}.
\]
This mismatch is especially important in the low-NFE regime, where each solver step is large and
early prediction errors can move the trajectory away from the forward-noised data manifold.

Under squared clean-prediction loss, the ideal training-time predictor at timestep \(t_i\) is the
conditional mean
\[
m_i(x) = \mathbb{E}_{q}[x_0 \mid x_{t_i}=x].
\]
The denoiser \(D_\theta\) is trained to approximate this target on samples from \(q_{t_i}\).
At inference time, however, it is queried on samples from \(p_i\). When \(p_i\) contains states
that are less data-like than the training marginal, the clean prediction \(\hat{x}_{0,i}\) can be
biased or unstable. This motivates using information from a nearby later state on the same deterministic
trajectory.

\paragraph{Pixel-Space and Latent-Space Usage.}
The same notation applies to both pixel-space and latent-space diffusion models.
For pixel-space models, \(z_i\) and \(\hat{x}_{0,i}\) are image-space tensors.
For latent diffusion models, they are latent tensors, and decoding is applied only after sampling
for evaluation or visualization. In both cases, PEEK modifies the predicted clean sample in the space
where the deterministic solver operates.

\section{Further Implementation Details}
\label{app:implementation}

\paragraph{Solvers and Timestep Grids.}
Most reported \method\ runs use deterministic first-order solver wrappers to keep the comparison
variables controlled. The DPM-Solver++ rows use the dpmsolver++ algorithm type in the scheduler
implementation \citep{lu2022dpmsolverpp}; the MS-COCO higher-order check additionally uses the order-3
DPM-Solver++ setting. DEIS rows use the DEIS update rule \citep{zhang2023deis} but,
unless explicitly marked otherwise, reuse the DPM-Solver++ binary interpolation schedule for the
same dataset and base step count. This makes the DEIS result a transfer test of the learned
correction positions while preserving the same $N+K$ denoiser-evaluation budget as its NFE-matched DEIS baseline.

\paragraph{Models, Resolution, and Conditioning.}
CIFAR-10 Model A uses the google/ddpm-cifar10-32 checkpoint evaluated at $32^2$ resolution,
and CIFAR-10 Model B uses the DanielBairamian/ddpm-cifar10-32-ema checkpoint for
the second CIFAR study and \methodg\ CIFAR rows. Both checkpoints expose the
same UNet2DModel macro-architecture in Diffusers: 35.75M parameters, block widths
$(128,256,256,256)$, two residual layers per block, and one attention block at the
16-by-16 scale.
The basic LSUN model uses google/ddpm-ema-bedroom-256. The EDM table additionally uses the
ImageNet $64^2$ conditional EDM checkpoint and the EDM LSUN model checkpoint
at $256^2$ \citep{karras2022edm}. For the EDM LSUN model, we include an iPNDM
baseline \citep{liu2022pndm}. ImageNet uses facebook/DiT-XL-2-256 in latent space
with classifier-free guidance scale 4.0. MS-COCO uses a Stable Diffusion v1.5 checkpoint
fine-tuned/evaluated with guidance scale 7.5, discrete latent-diffusion timesteps, and $512 \times 512$
decoding before FID evaluation. The latent-space experiments interpolate the predicted clean latent,
not decoded RGB images.

\paragraph{Distributed Evaluation.}
The pixel-space and latent-space evaluations use multi-GPU H200 nodes with one
sample shard per process. The CIFAR-10 qualitative grid uses saved snapshots
from the 9-step proxy runs.

\paragraph{Interpolation in Pixel and Latent Spaces.}
For pixel-space DDPM models, the clean prediction in the main-paper PEEK interpolation equation is an image-space
prediction. For latent diffusion models, the same operation is applied to the predicted latent clean
sample. FID and qualitative visualizations are computed after the standard decoder is applied.

\paragraph{Candidate Gates.}
Vanilla \method\ uses binary interpolation gates $\{0,1\}$. \methodg\ uses the
direction-space candidate set $\{0,0.25,0.5,0.75,1\}$.

\paragraph{NFE-Matched Baseline.}
If an $N$-step \method\ schedule uses $K$ nonzero look-ahead positions, then the deployment run
uses $N+K$ denoiser evaluations. The NFE-matched baseline is the same deterministic solver run with
$N+K$ ordinary solver steps and no look-ahead interpolation.

\paragraph{Restart-Style Repeat Baseline.}
The Restart-style baseline follows the same NFE budget as
the corresponding \method\ run. After each selected main step,
it forwards the current state from the next
lower-noise timestep back to the previous
higher-noise timestep by adding fresh Gaussian noise,
then runs one deterministic solver step back to the lower-noise timestep.
The selected repeat positions are the nonzero entries of the \method\ schedule,
so a schedule with $K$ nonzero entries also spends $K$ extra denoiser evaluations \citep{xu2023restart}.

\paragraph{Evaluation Artifacts.}
Final FID values are measured on held-out samples distinct from those used for
validation search. We use the StyleGAN Inception feature extractor for FID statistics,
with 2048-dimensional features for reported final FID except where explicitly marked as a
256-dimensional proxy. CIFAR-10, ImageNet, and MS-COCO schedules are selected
with lightweight proxy runs and then rerun with the fixed schedules under the
standard final-reporting protocol. Unless otherwise stated, qualitative \method\ and \methodg\
figures use DPM-Solver++ schedules and samples.

\paragraph{Selected Schedules.}
Table~\ref{tab:app_schedules} lists the schedules used for the main final evaluations.
Each entry has one fewer interpolation gate than solver steps because the final denoising step has
no later timestep to peek at. The DEIS final evaluation on the basic LSUN model uses the
corresponding DPM-Solver++ schedule as a transfer test; the EDM DEIS and iPNDM
rows likewise reuse the corresponding DPM-Solver++ schedule. All other rows use the schedule searched with
the same sampler as the final evaluation.

\begin{table*}[t]
  \centering
  \small
  \setlength{\tabcolsep}{3pt}
  \begin{tabular}{@{}>{\raggedright\arraybackslash}p{0.36\linewidth}>{\raggedright\arraybackslash}p{0.18\linewidth}ccp{0.19\linewidth}@{}}
    \toprule
    Dataset & Evaluation sampler & Base NFE & Total NFE & $\boldsymbol{\gamma}$ \\
    \midrule
    \multicolumn{5}{l}{\method} \\
    Unconditional Pixel-Space Generation on CIFAR-10 & DPM-Solver++ & 9 & 17 & $(1,1,1,1,1,1,1,1)$ \\
    Unconditional Pixel-Space Generation on CIFAR-10 (EMA, \method) & DPM-Solver++ & 9 & 11 & $(0,0,0,0,0,0,1,1)$ \\
    Unconditional Pixel-Space Generation on CIFAR-10 (EMA, \method) & DPM-Solver++ & 20 & 23 & $\begin{array}[t]{@{}l@{}}(0,0,0,0,0,0,0,\\0,0,0,0,0,0,0,\\0,0,1,1,1)\end{array}$ \\
    Unconditional Pixel-Space Generation on Basic LSUN Bedroom Model & DPM-Solver++ / DEIS & 6 & 10 & $(0,1,1,1,1)$ \\
    Class-Conditional Latent Generation on ImageNet $256 \times 256$ & DPM-Solver++ & 9 & 13 & $(0,0,0,0,1,1,1,1)$ \\
    Text-Conditional Latent Generation on MS-COCO & DPM-Solver++ (order 1 / 3) & 9 & 12 & $(0,0,0,0,0,1,1,1)$ \\
    Conditional EDM Generation on ImageNet $64 \times 64$ & DPM-Solver++ / DEIS & 6 & 9 & $(1,1,1,0,0)$ \\
    Unconditional EDM Generation on EDM LSUN Model & DPM-Solver++ / DEIS / iPNDM & 5 & 8 & $(1,1,1,0)$ \\
    \midrule
    \multicolumn{5}{l}{\methodg} \\
    Unconditional Pixel-Space Generation on CIFAR-10 (EMA, \methodg) & DPM-Solver++ & 9 & 13 & $(1,0,0,0,0,1,1,1)$ \\
    Unconditional Pixel-Space Generation on Basic LSUN Model (\methodg) & DPM-Solver++ & 6 & 10 & $(0,0.5,1,0.5,0.5)$ \\
    Class-Conditional Latent Generation on ImageNet $256 \times 256$ (\methodg) & DPM-Solver++ & 9 & 13 & $\begin{array}[t]{@{}l@{}}(0,0,0,0,0.5,\\0.5,0.25,1)\end{array}$ \\
    Text-Conditional Latent Generation on MS-COCO (\methodg) & DPM-Solver++ (order 1) & 9 & 13 & $\begin{array}[t]{@{}l@{}}(0,0,0,0,1,\\1,1,0.25)\end{array}$ \\
    \bottomrule
  \end{tabular}
  \caption{Interpolation schedules used in the reported final evaluations. NFE counts the ordinary solver calls plus nonzero look-ahead calls. Vanilla \method\ uses binary gates, while \methodg\ searches over $\{0,0.25,0.5,0.75,1\}$.}
  \label{tab:app_schedules}
\end{table*}

\paragraph{Calibration Cost.}
Schedule calibration uses proxy metrics. On CIFAR-10, we use a
1,000-sample, 256-dimensional proxy for the
DPM-Solver++ schedule sweep, then rerun the fixed schedules with the standard
final-reporting protocol. The 9-step schedule search took 20.4
seconds, and the 20-step schedule search took 32.5 seconds; both
used about 2.2 GB peak CPU resident set size and 8.7 GB peak CUDA reserved
memory. For the latent ImageNet and MS-COCO runs, the main-paper additional-datasets table reports
the proxy-search time and peak CUDA memory for \method\ before the selected
schedules are rerun with the standard final evaluation metric.
 Once calibration is complete, the schedule is fixed and
 no search is run at deployment, so inference speed is determined
 only by the selected NFE budget.

\section{Extended Related Work}
\label{app:extended_related_work}

\paragraph{Training and Distillation for Fast Generation.}
One way to reduce few-step sampling error is to train or distill a model specifically for
short trajectories. Progressive distillation compresses a many-step diffusion sampler into a smaller
number of student steps \citep{salimans2022progressive}, while consistency models learn mappings that
support direct or few-step generation \citep{song2023consistency}. Recent lightweight variants optimize
a small sampler or conditioning component instead of the whole denoiser: GAS distills and adversarially
trains a generalized ODE solver \citep{oganov2025gas}, and MTEO freezes the pretrained backbone while
optimizing step-wise, layer-wise time embeddings. These approaches can deliver strong low-NFE deployment
with little or no per-step overhead after training \citep{cai2026mteo}. \method\ addresses a different
operating point: it is training-free and can be applied to an existing sampler immediately, at the cost
of selected extra denoiser calls during inference.

\paragraph{Sampler and Schedule Design.}
DDIM introduced deterministic non-Markovian sampling paths that can use fewer denoising steps than ancestral sampling \citep{song2021ddim}. PNDM, DPM-Solver, DPM-Solver++, DEIS, UniPC, GENIE, AMED-Solver, and DPM-Solver-v3 further improve low-NFE performance through ODE-aware integration, multistep history, exponential-integration formulas, learned directions, solver statistics, or predictor-corrector design \citep{liu2022pndm,lu2022dpmsolver,lu2022dpmsolverpp,zhang2023deis,zhao2023unipc,dockhorn2022genie,zhou2023fast,zheng2023dpmsolverv3}. Heun's method is a representative predictor-corrector update: it forms a predicted next state and combines the current and predicted ODE derivatives, whereas \method\ uses the predicted state only to obtain a clean-sample estimate that can interpolate the current clean prediction before the next noisy state is formed. EDM and schedule-alignment work show that noise parameterization, timestep placement, and architecture allocation are also central to sample quality \citep{karras2022edm,li2023autodiffusion,sabour2024align,chen2024trajectory,pei2025optimalsteps}. SDM formalizes an adaptive sampling design that uses low-order solvers in early high-noise stages and progressively deploys higher-order solvers later \citep{jo2026sdm}. ERK-Guid instead exploits embedded Runge-Kutta local-error estimates as a guidance signal, so its mechanism is specific to embedded/higher-order ODE-solver structure \citep{kong2026erkguid}. \method\ is orthogonal to these advances: it uses the same timestep grid at deployment and changes the denoiser-derived clean estimate inside selected deterministic steps.

\paragraph{Inference-Time Correction.}
Restart sampling alternates a backward ODE path with forward noising segments,
using stochasticity to contract accumulated sampling error \citep{xu2023restart}.
Among prior training-free inference-time methods, this is the closest in spirit because it
spends extra denoiser evaluations on correction rather than only densifying an ordinary trajectory.
The mechanism is different: Restart exposes the sampler to a fresh-noise sub-trajectory and requires
interval boundaries, restart counts, and sub-trajectory lengths; \method\ uses a deterministic provisional
next state only to obtain a one-step-later clean prediction, then applies the correction to the current step.
Recent adaptive-stochasticity plugins make a different tradeoff. AdaSDE learns a per-step
stochastic coefficient through lightweight distillation, dynamically regulating correction
strength while remaining compatible with existing solvers \citep{wang2025adasde}.
\method\ uses no distillation and no injected stochasticity; its correction is deterministic
once the schedule is fixed, so it can be layered conceptually with stochastic or learned
correction methods but occupies the simpler training-free setting. DiffRS is also post-hoc,
but uses a learned time-dependent discriminator to accept or reject intermediate samples and
therefore refines different samples with different effort \citep{na2024diffrs}. DiffRS is attractive
when sample-adaptive quality filtering and auxiliary discriminator training are acceptable, whereas
\method\ is preferable when a fixed deployment NFE budget, no extra learned model, and a simple
deterministic trajectory are desired. SE2P is another relevant inference-time correction: it uses
two parallel samplers at successive times and integrates information from the later-step processor
into the earlier one without adding extra model calls per processor \citep{cisneros2025tango}.
This is conceptually close to \method{}'s use of a later-step state as a correction signal, but
the computational assumptions differ. SE2P relies on parallel processors, whereas \method\ uses explicit
extra look-ahead evaluations within one sampler and reports comparisons against NFE-matched ordinary
trajectories. Related analyses of diffusion trajectories and schedule flaws emphasize that sampler
states encountered during inference can differ from idealized training or terminal distributions
\citep{lin2023flawed,chen2023geometric}. \method\ targets this issue without injecting fresh noise,
 changing the timestep grid, or deriving a new solver: the correction comes from the model's own
 next-step clean prediction along the current deterministic trajectory and is selected with an
 automated metric-driven search.

\section{Pseudocode}
\label{app:pseudocode}

\paragraph{\method\ Sampling.}
Given timesteps $t_0,\ldots,t_{N-1}$, initial state $z_0$, and schedule $\boldsymbol{\gamma}$:
\begin{enumerate}
\item For $i=0,\ldots,N-2$, compute the ordinary prediction $\hat{x}_{0,i}^{\mathrm{base}}=D_\theta(z_i,t_i)$.
\item If $\gamma_i=0$, set $z_{i+1}=\Phi_i(z_i,\hat{x}_{0,i}^{\mathrm{base}})$ and continue.
\item Otherwise compute the provisional state $\tilde{z}_{i+1}=\Phi_i(z_i,\hat{x}_{0,i}^{\mathrm{base}})$.
\item Compute the look-ahead prediction $\hat{x}_{0,i}^{\mathrm{look}}=D_\theta(\tilde{z}_{i+1},t_{i+1})$.
\item Select $\hat{x}_{0,i}^{\mathrm{sel}}=\hat{x}_{0,i}^{\mathrm{look}}$.
\item Take the actual step $z_{i+1}=\Phi_i(z_i,\hat{x}_{0,i}^{\mathrm{sel}})$.
\item Run the final ordinary solver step at $t_{N-1}$.
\end{enumerate}

\paragraph{\methodg\ Sampling.}
\methodg\ follows the same provisional look-ahead procedure, but interpolates the
solver direction rather than the clean prediction. At step \(i\), compute the
ordinary direction \(d_i^{\mathrm{base}}\). If \(\gamma_i=0\), take the ordinary
update without a look-ahead query. Otherwise, form \(\tilde{z}_{i+1}\), query
the denoiser at \((\tilde{z}_{i+1},t_{i+1})\), convert the result to
\(d_i^{\mathrm{look}}\), and use
\((1-\gamma_i)d_i^{\mathrm{base}}+\gamma_i d_i^{\mathrm{look}}\) inside the
ordinary DPM-Solver update for the current interval. The final step remains an
ordinary solver step because no later timestep is available.

\paragraph{DPM-Solver with \methodg.}
For a DPM-Solver-style sampler, let \(U_i(z_i,d_i)\) denote the ordinary solver
update over interval \([t_i,t_{i+1}]\) after the denoiser output has been converted
to the solver direction \(d_i\). Given a direction schedule \(\boldsymbol{\gamma}\):
\begin{enumerate}
\item For \(i=0,\ldots,N-2\), evaluate the denoiser at the current state and form
the ordinary direction \(d_i^{\mathrm{base}}=G_i(z_i,D_\theta(z_i,t_i))\).
\item If \(\gamma_i=0\), set
\(z_{i+1}=U_i(z_i,d_i^{\mathrm{base}})\) and continue.
\item Take the provisional DPM-Solver step
\(\tilde{z}_{i+1}=U_i(z_i,d_i^{\mathrm{base}})\).
\item Query the denoiser at the provisional next state,
\(\hat{x}_{0,i}^{\mathrm{look}}=D_\theta(\tilde{z}_{i+1},t_{i+1})\), and convert it
to a look-ahead direction \(d_i^{\mathrm{look}}=G_{i+1}(\tilde{z}_{i+1},\hat{x}_{0,i}^{\mathrm{look}})\).
\item Blend the direction according to the selected gate:
\[
d_i^{\mathrm{sel}}=(1-\gamma_i)d_i^{\mathrm{base}}+\gamma_i d_i^{\mathrm{look}}.
\]
\item Apply the ordinary DPM-Solver update with the original interval coefficients:
\[
z_{i+1}=U_i(z_i,d_i^{\mathrm{sel}}).
\]
\item At \(t_{N-1}\), run the final ordinary DPM-Solver step, because there is no
later timestep from which to obtain a look-ahead direction.
\end{enumerate}

\paragraph{Greedy Schedule Search.}
For each peekable position $i$, the search reuses the selected prefix state, evaluates candidate interpolation gates on the remaining suffix, compares the best candidate to the NFE-matched ordinary baseline, and stores the selected prefix state for position $i+1$. Vanilla \method\ uses $\mathcal{A}=\{0,1\}$, while \methodg\ uses $\mathcal{A}=\{0,0.25,0.5,0.75,1\}$. The selected schedule is then used unchanged for final evaluation.

\section{Cache Derivation}
\label{app:cache_details}

\paragraph{Worst-Case Count.}
We count denoiser queries and index the $N-1$ peekable positions by
$i=0,\ldots,N-2$. In the worst case, every previous gate is nonzero, so
the NFE-matched budget at position $i$ is $N+i+1$. Without caching, the
initial current schedule costs $N$ queries. The ordinary baseline and the
\method\ candidate are then both rerun in full at each budget from $N+1$
through $2N-1$. Thus
\begin{equation}
  C_n = N+2\sum_{i=0}^{N-2}(N+i+1).
\end{equation}
Changing the summation index gives
\begin{equation}
  C_n = N+2\sum_{m=N+1}^{2N-1}m = 3N^2-2N.
\end{equation}

The two caches change the count as follows. The initial partial schedule is
evaluated once at a cost of $N$ queries, and its metric supplies the zero-gate
candidate at the first position. The first ordinary baseline costs $N+1$
queries. At each later position $i=1,\ldots,N-2$, baseline rebasing maps the
cached pre-final state to the new grid, leaving only the new final-step query.
Consequently,
\begin{equation}
  C_b
  = N+1+\sum_{i=1}^{N-2}1
  = 2N-1.
\end{equation}
For the \method\ candidate at position $i$, the selected prefix contains
all earlier base and look-ahead results. Only one new base query, one new
look-ahead query, and the $N-1-i$ ordinary queries in the remaining suffix
must be evaluated. Therefore,
\begin{equation}
  C_p = \sum_{i=0}^{N-2}\left[1+1+(N-1-i)\right].
\end{equation}
Reversing the suffix index gives
\begin{equation}
  C_p = 2(N-1)+\sum_{r=1}^{N-1}r
  = \frac{N^2+3N-4}{2}.
\end{equation}
Including the initial partial schedule, the cached search cost is
\begin{equation}
  C_c = N+C_b+C_p = \frac{N^2+9N-6}{2}.
\end{equation}
For the paper's $N=9$ search, the naive count is
$9+2(10+\cdots+17)=225$. The initial partial schedule costs 9, the cached baseline costs
$10+7\cdot1=17$, and the cached \method\ candidates cost
$2\cdot8+(8+\cdots+1)=52$, for a total of 78. The reduction is
$(225-78)/225=65.3\%$. These sums count denoiser queries; cached base and
look-ahead results incur no new query. Additional nonzero gate candidates
add their corresponding \method\ candidate work.

\section{Example Search Log}
\label{app:search_log}

\small
\begin{quote}
step 0: NFE-matched baseline FID 3.0802; current \method\ FID 3.2585; candidate $\gamma=1$ gives 3.2459; selected $\gamma=0$.\\
step 5: NFE-matched baseline FID 3.0802; current \method\ FID 3.2585; candidate $\gamma=1$ gives 3.0634; selected $\gamma=1$.\\
step 7: NFE-matched baseline FID 2.9720; current \method\ FID 2.9003; candidate $\gamma=1$ gives 2.7922; selected $\gamma=1$.
\end{quote}
\normalsize

\section{Reproducibility and Responsible Use}
\label{app:reproducibility}

All reported results use pretrained image generators and
image benchmarks. The method does not introduce
new training data or a new model checkpoint; it is an
inference procedure plus a schedule-calibration routine.
Code and reproduction materials are available from the project website at
\url{https://dachengruoque.github.io/PEEK-Video/}, including the schedules,
commands, and evaluation utilities needed to reproduce the reported
sampling results.
  Image generation systems can be misused to synthesize
   deceptive or harmful visual content. \method\ can
    improve generation quality, so releases should
     follow the same safeguards and usage restrictions
     as the underlying generative models.

\paragraph{Existing Assets.}
The experiments use benchmarks and pretrained generators: CIFAR-10 \citep{krizhevsky2009learning}, LSUN Bedroom \citep{yu2015lsun}, ImageNet \citep{deng2009imagenet}, MS-COCO \citep{lin2014microsoft}, DDPM-family checkpoints \citep{ho2020ddpm}, DiT \citep{peebles2023scalable}, Stable Diffusion \citep{rombach2022latent}, and solver implementations following DPM-Solver++, DEIS, UniPC, and fast-sampler codebases \citep{lu2022dpmsolverpp,zhang2023deis,zhao2023unipc,zhou2023fast,chen2024trajectory}.

\section{Additional Visualizations}
\label{app:additional_visualizations}
\begin{figure*}[p]
  \centering
  \includegraphics[width=\textwidth,height=0.78\textheight,keepaspectratio]{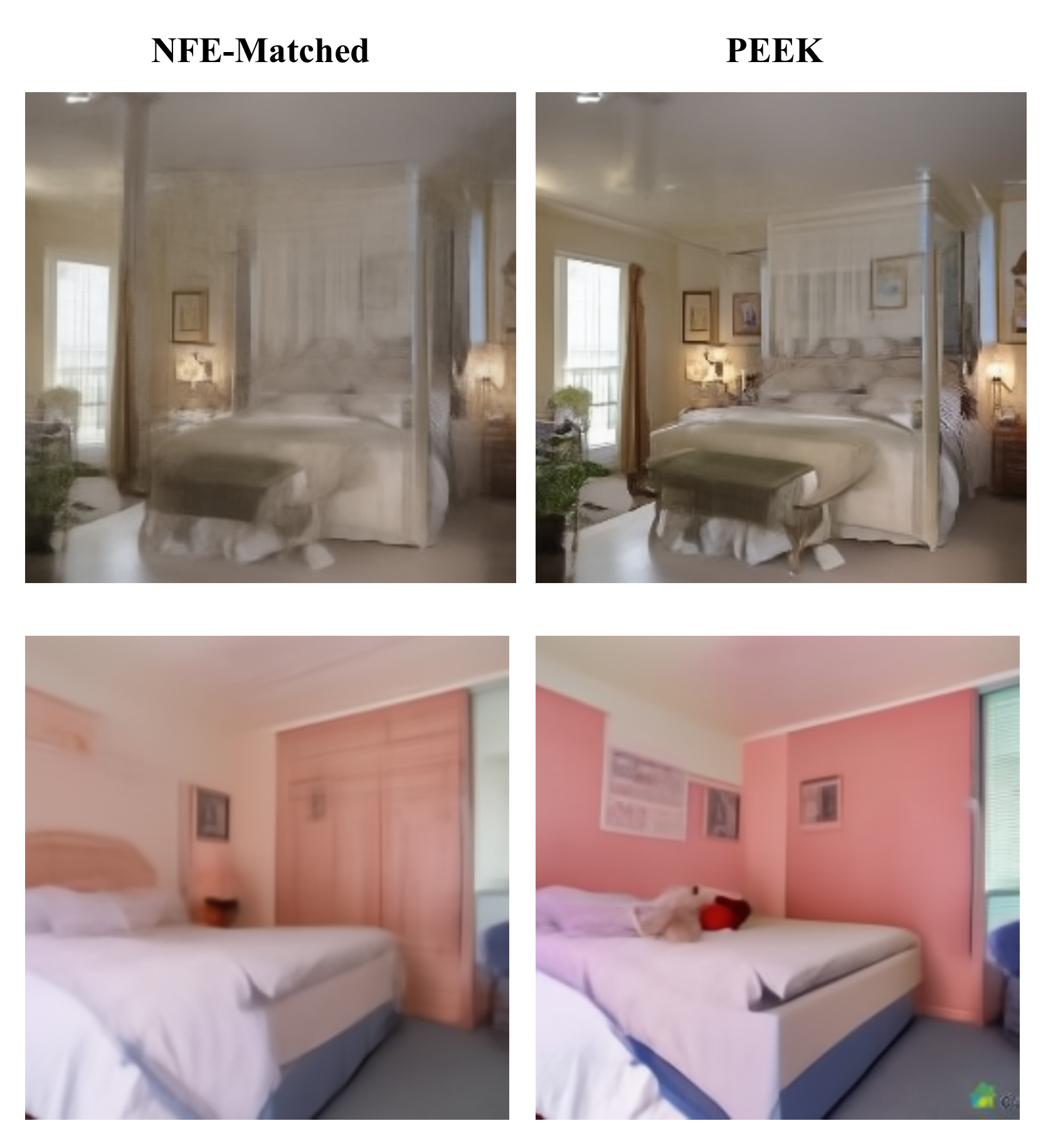}
  \caption{Basic LSUN Bedroom model \method\ Qualitative Comparison I. Samples are generated by
  \method\ with
  the google/ddpm-ema-bedroom-256 pixel-space DDPM checkpoint. Best viewed in color.}
  \label{fig:app2}
\end{figure*}

\begin{figure*}[p]
  \centering
  \includegraphics[width=\textwidth,height=0.78\textheight,keepaspectratio]{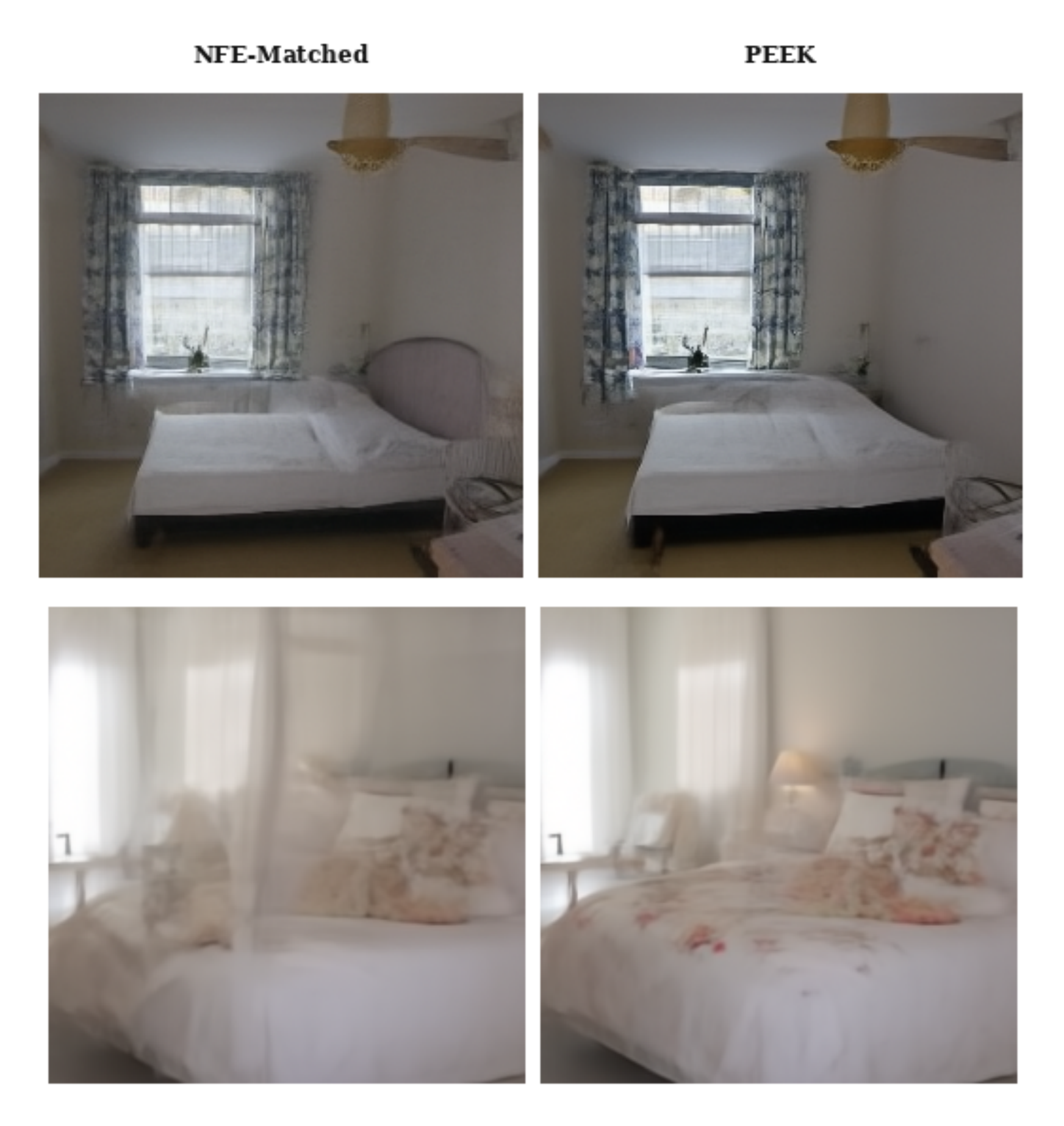}
  \caption{Basic LSUN Bedroom model \method\ Qualitative Comparison II. Samples are generated by
  \method\ with
  the google/ddpm-ema-bedroom-256 pixel-space DDPM checkpoint. Best viewed in color.}
  \label{fig:app5}
\end{figure*}

\begin{figure*}[p]
  \centering
  \includegraphics[width=\textwidth,height=0.78\textheight,keepaspectratio]{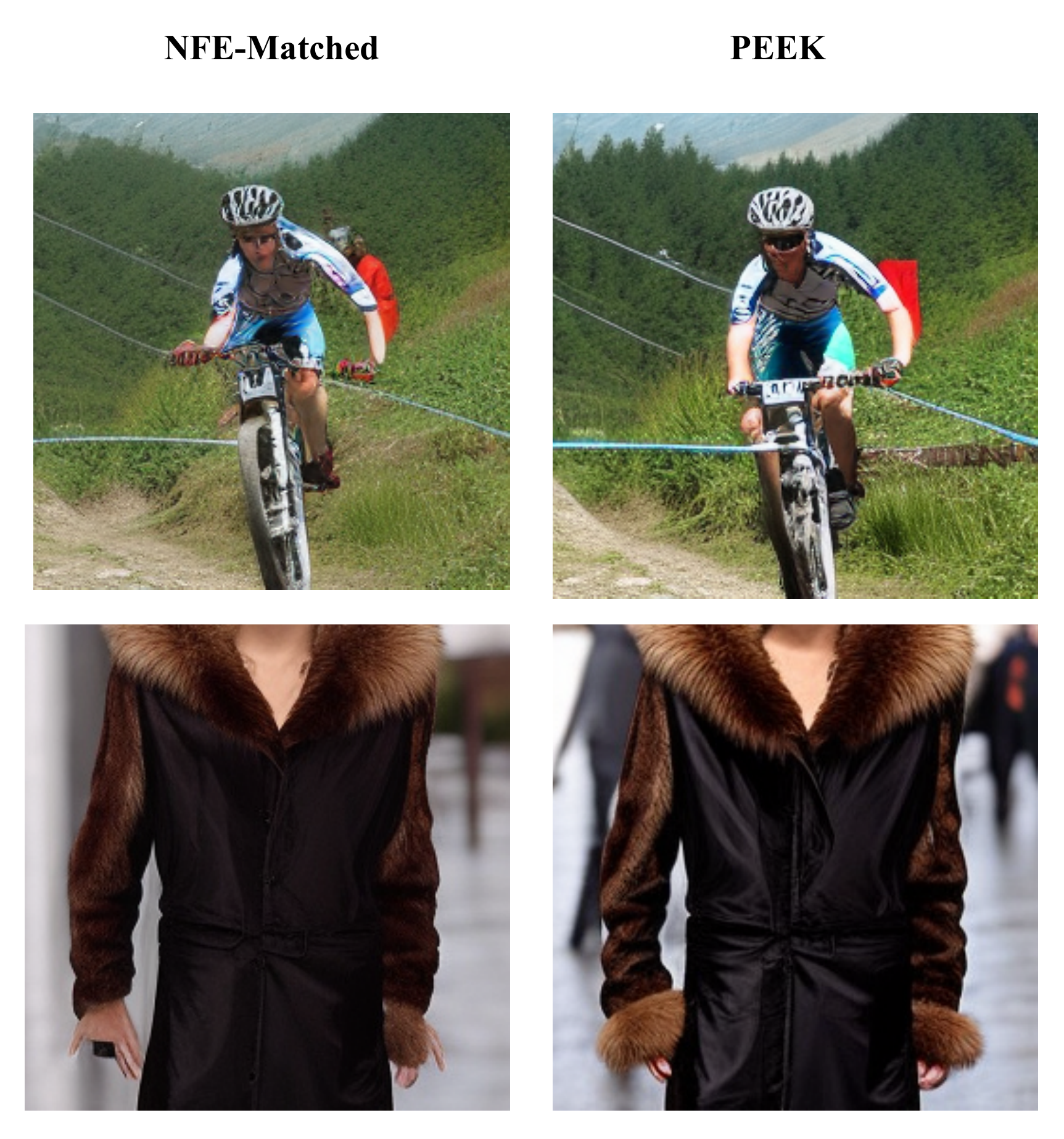}
  \caption{ImageNet \method\ Qualitative Comparison. Samples are generated by
  \method\ with the
  facebook/DiT-XL-2-256 class-conditional latent diffusion transformer with
  classifier-free guidance. Best viewed in color.}
  \label{fig:app6}
\end{figure*}

\begin{figure*}[p]
  \centering
  \includegraphics[width=\textwidth,height=0.78\textheight,keepaspectratio]{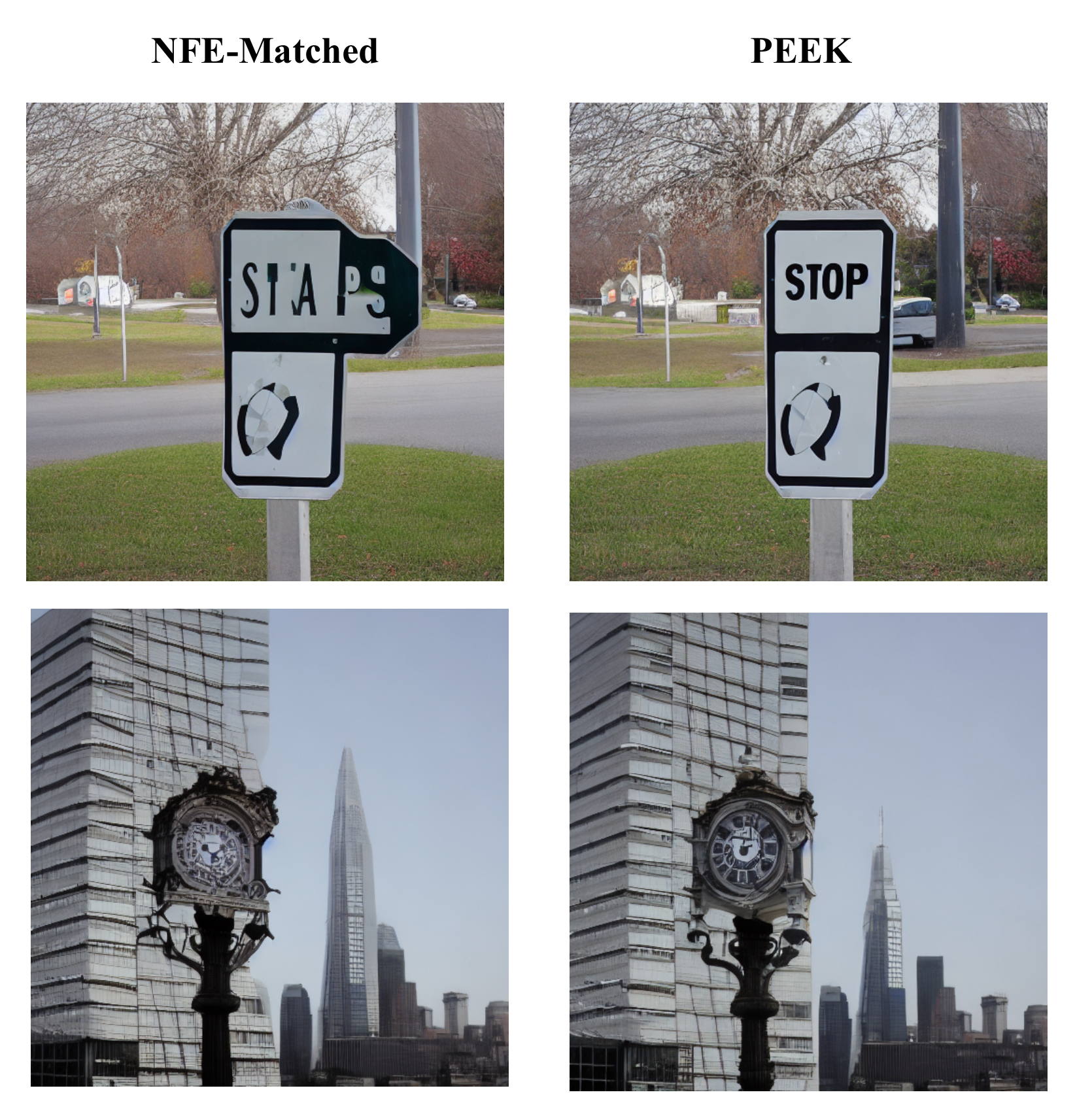}
  \caption{MS-COCO \method\ Qualitative Comparison I. Samples are generated by
  \method\ with the
  Stable Diffusion v1.5 latent text-to-image checkpoint. Best viewed in color.}
  \label{fig:app3}
\end{figure*}

\begin{figure*}[p]
  \centering
  \includegraphics[width=\textwidth,height=0.78\textheight,keepaspectratio]{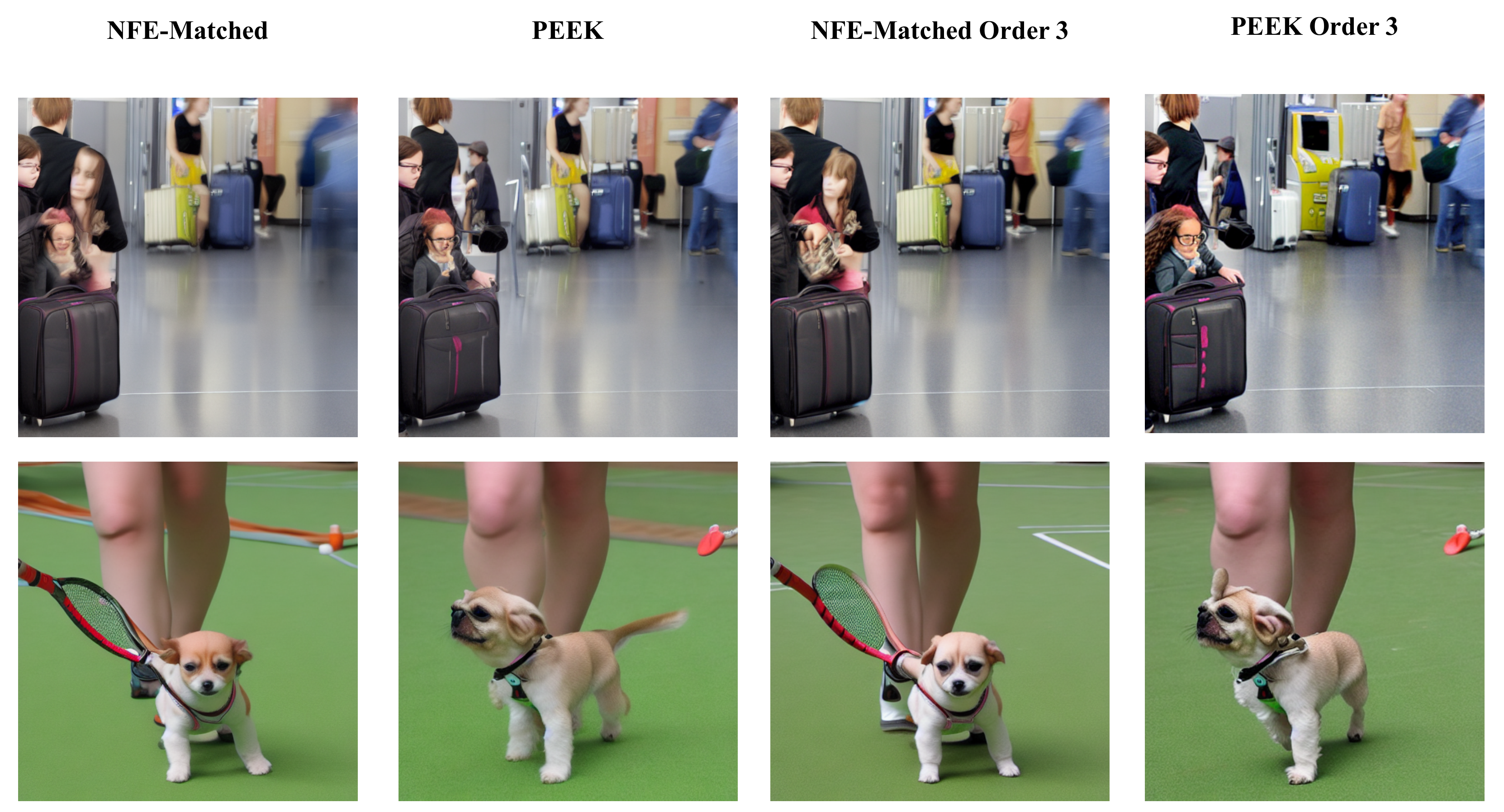}
  \caption{MS-COCO \method\ Qualitative Comparison II. Samples are generated by
  \method\ with the
  Stable Diffusion v1.5 latent text-to-image checkpoint. Best viewed in color.}
  \label{fig:app4}
\end{figure*}

\begin{figure*}[p]
  \centering
  \includegraphics[width=\textwidth,height=0.78\textheight,keepaspectratio]{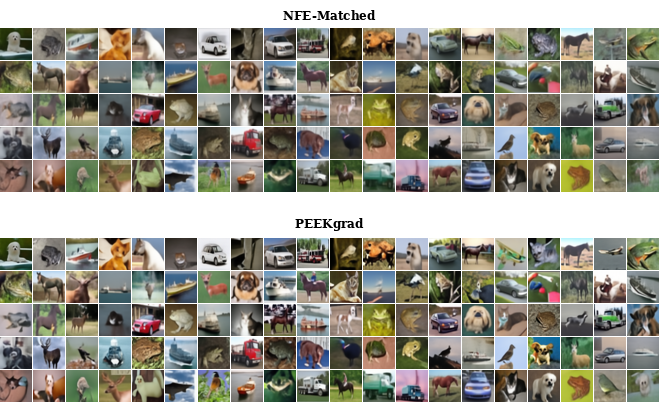}
  \caption{CIFAR-10 \methodg\ Qualitative Comparison. Each method block contains five
  rows with 20 samples per row. \methodg\ is compared against
  the NFE-matched DPM-Solver++ baseline under the same denoiser-evaluation budget. Best viewed in color.}
  \label{fig:cifar_qual_compare}
\end{figure*}

\begin{figure*}[p]
  \centering
  \includegraphics[width=\textwidth,height=0.78\textheight,keepaspectratio]{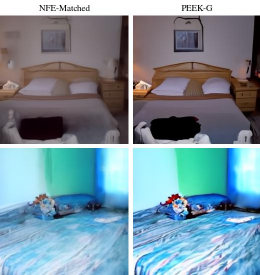}
  \caption{Basic LSUN Bedroom model \methodg\ Qualitative Comparison I. Samples are generated by
  \methodg\ with
  the google/ddpm-ema-bedroom-256 pixel-space DDPM checkpoint. Best viewed in color.}
  \label{fig:lsun_peekgrad_good_1}
\end{figure*}

\begin{figure*}[p]
  \centering
  \includegraphics[width=\textwidth,height=0.78\textheight,keepaspectratio]{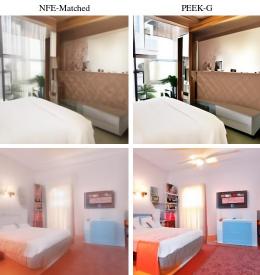}
  \caption{Basic LSUN Bedroom model \methodg\ Qualitative Comparison II. Samples are generated by
  \methodg\ with
  the google/ddpm-ema-bedroom-256 pixel-space DDPM checkpoint. Best viewed in color.}
  \label{fig:lsun_peekgrad_good_2}
\end{figure*}

\begin{figure*}[p]
  \centering
  \includegraphics[width=\textwidth,height=0.78\textheight,keepaspectratio]{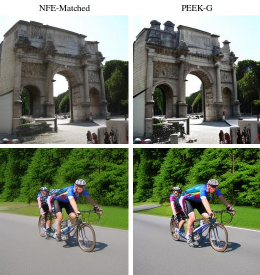}
  \caption{ImageNet \methodg\ Qualitative Comparison I. Samples are generated by
  \methodg\ with the
  facebook/DiT-XL-2-256 class-conditional latent diffusion transformer with
  classifier-free guidance. Best viewed in color.}
  \label{fig:imagenet_peekgrad_good_1}
\end{figure*}

\begin{figure*}[p]
  \centering
  \includegraphics[width=\textwidth,height=0.78\textheight,keepaspectratio]{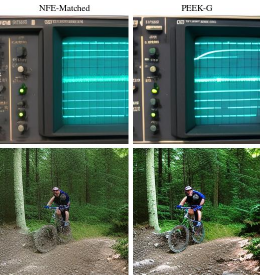}
  \caption{ImageNet \methodg\ Qualitative Comparison II. Samples are generated by
  \methodg\ with the
  facebook/DiT-XL-2-256 class-conditional latent diffusion transformer with
  classifier-free guidance. Best viewed in color.}
  \label{fig:imagenet_peekgrad_good_2}
\end{figure*}

\begin{figure*}[p]
  \centering
  \includegraphics[width=\textwidth,height=0.78\textheight,keepaspectratio]{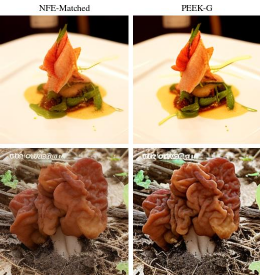}
  \caption{ImageNet \methodg\ Qualitative Comparison III. Samples are generated by
  \methodg\ with the
  facebook/DiT-XL-2-256 class-conditional latent diffusion transformer with
  classifier-free guidance. Best viewed in color.}
  \label{fig:imagenet_peekgrad_good_3}
\end{figure*}

\begin{figure*}[p]
  \centering
  \includegraphics[width=\textwidth,height=0.78\textheight,keepaspectratio]{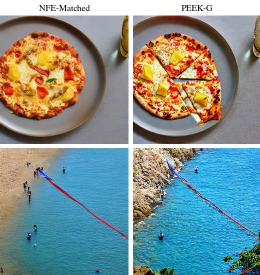}
  \caption{MS-COCO \methodg\ Qualitative Comparison I. Samples are generated by
  \methodg\ with the
  Stable Diffusion v1.5 latent text-to-image checkpoint. Best viewed in color.}
  \label{fig:coco_peekgrad_good_2}
\end{figure*}

\begin{figure*}[p]
  \centering
  \includegraphics[width=\textwidth,height=0.78\textheight,keepaspectratio]{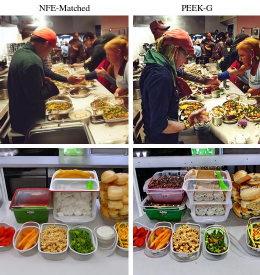}
  \caption{MS-COCO \methodg\ Qualitative Comparison II. Samples are generated by
  \methodg\ with the
  Stable Diffusion v1.5 latent text-to-image checkpoint. Best viewed in color.}
  \label{fig:coco_peekgrad_good_4}
\end{figure*}

\begin{figure*}[p]
  \centering
  \includegraphics[width=\textwidth,height=0.78\textheight,keepaspectratio]{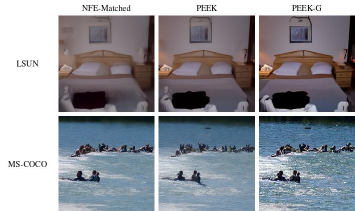}
  \caption{\method\ and \methodg\ Qualitative Contrast. Rows show
  the basic LSUN Bedroom model and MS-COCO. Columns show the NFE-matched DPM-Solver++
  baseline, vanilla \method, and \methodg. Best viewed in color.}
  \label{fig:peek_peekgrad_qualitative}
\end{figure*}

Figures~\ref{fig:app2}-\ref{fig:app4} provide additional \method\ samples,
and Figures~\ref{fig:cifar_qual_compare}-\ref{fig:peek_peekgrad_qualitative}
provide \methodg\ samples and direct qualitative contrasts.

\clearpage

\clearpage

\clearpage

\clearpage

\clearpage

\clearpage

\clearpage

\clearpage

\clearpage

\clearpage

\clearpage

\clearpage

\clearpage

\clearpage

\end{document}